\documentclass[runningheads]{llncs}

\usepackage[mobile]{eccv}

\usepackage{eccvabbrv}

\usepackage{graphicx}
\usepackage{booktabs}
\usepackage[shortcuts,acronym]{glossaries}
\usepackage{multirow} 

\usepackage[accsupp]{axessibility}  %
\usepackage{xcolor}
\definecolor{DarkGreen}{RGB}{0,150,0} \definecolor{DarkRed}{RGB}{180,0,0} \definecolor{DarkGray}{gray}{0.5}

\usepackage[pagebackref,breaklinks,colorlinks,citecolor=eccvblue]{hyperref}
\usepackage{orcidlink}

\newacronym{bev}{BEV}{Bird’s-Eye View}
\newacronym{lidar}{LiDAR}{Light Detection and Ranging}
\newacronym{uav}{UAV}{Unmanned Aerial Vehicle}
\newacronym{nuscenes}{nuScenes}{nuScenes Autonomous Driving Dataset}
\newacronym{mse}{MSE}{Mean Squared Error}
\newacronym{cnn}{CNN}{Convolutional Neural Network}
\newacronym{nds}{NDS}{nuScenes Detection Score}
\newacronym{hd}{HD}{High Definition}
\newacronym{gt}{GT}{Ground Truth}
\newacronym{map}{mAP}{mean Average Precision}
\newacronym{roi}{RoI}{Region of Interest}
\newacronym{cka}{CKA}{Centered Kernel Alignment}
\newacronym{cvs}{CVS}{Cross-View Supervision}

\begin{document}

\title{Learning Ego-Centric BEV Representations from a Perspective-Privileged View:
Cross-View Supervision for Online HD Map Construction} 

\titlerunning{Perspective-Privileged Cross-View Supervision}

\author{
Daniel Lengerer$^{1}$~\orcidlink{0009-0009-1600-6144}\thanks{Corresponding author. E-mail: Daniel.Lengerer@tha.de},
Mathias Pechinger$^{2}$~\orcidlink{0000-0003-2371-9870},\\
Klaus Bogenberger$^{2}$~\orcidlink{0000-0003-3868-9571},
Carsten Markgraf$^{1}$~\orcidlink{0000-0001-9447-2065}
}

\authorrunning{Lengerer et al.}

\institute{Technical University of Applied Sciences Augsburg, Germany \and
Technical University of Munich, Germany\\
}
\maketitle

\begin{abstract}
\Gls{bev} representations derived from multi-camera input have become a central interface for online \acrshort{hd} map construction. 
However, most approaches rely solely on ego-centric supervision, requiring large-scale scene structure to be inferred from incomplete observations, occlusions, and diminishing information density at long range, where perspective effects and spatial sparsity hinder consistent structural reasoning.
We introduce \gls{cvs}, a representation learning paradigm that transfers geometric and topological priors from an ego-aligned overhead perspective into camera-based \gls{bev} encoders. 
Rather than adding auxiliary semantic losses, \gls{cvs} aligns representations in a shared \gls{bev} feature space and distills globally consistent structural knowledge from a perspective-privileged teacher into the ego-centric backbone. 
This supervision enhances structural coherence without modifying the inference architecture or requiring overhead input at test time.
Experiments on nuScenes using ego-aligned aerial imagery from the AID4AD cross-view extension demonstrate consistent improvements over StreamMapNet while maintaining identical camera-only inference.
\gls{cvs} yields +3.9\,mAP in the standard $60\times30\,\mathrm{m}$ region and +9.9\,mAP in the extended $100\times50\,\mathrm{m}$ setting, corresponding to a 44\% relative gain at long range.
These results highlight perspective-privileged structural supervision as a promising training principle for improving \gls{bev} representation learning in \acrshort{hd} map construction. The project repository is available at \\ \url{https://github.com/DriverlessMobility/CrossViewSupervision}.
\end{abstract}

\section{Introduction}

\begin{figure}[t]
  \centering
  \setlength{\tabcolsep}{2pt}
  \renewcommand{\arraystretch}{1.1}

  \begin{tabular}{cc}
    \textbf{Baseline} & \textbf{Ours} \\

    \begin{subfigure}{0.42\linewidth}
      \centering
      \includegraphics[width=\linewidth]{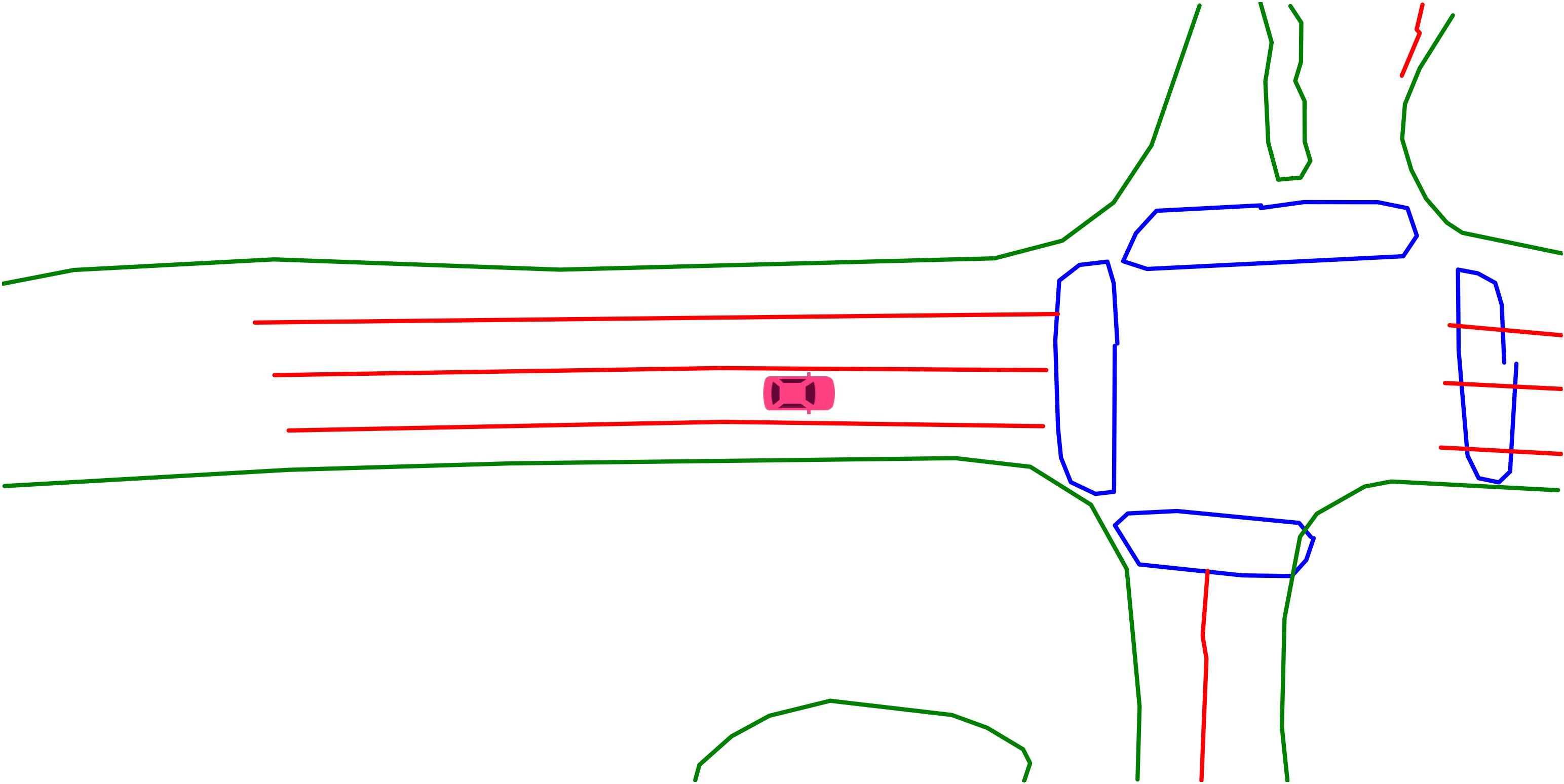}
    \end{subfigure} &
    \begin{subfigure}{0.42\linewidth}
      \centering
      \includegraphics[width=\linewidth]{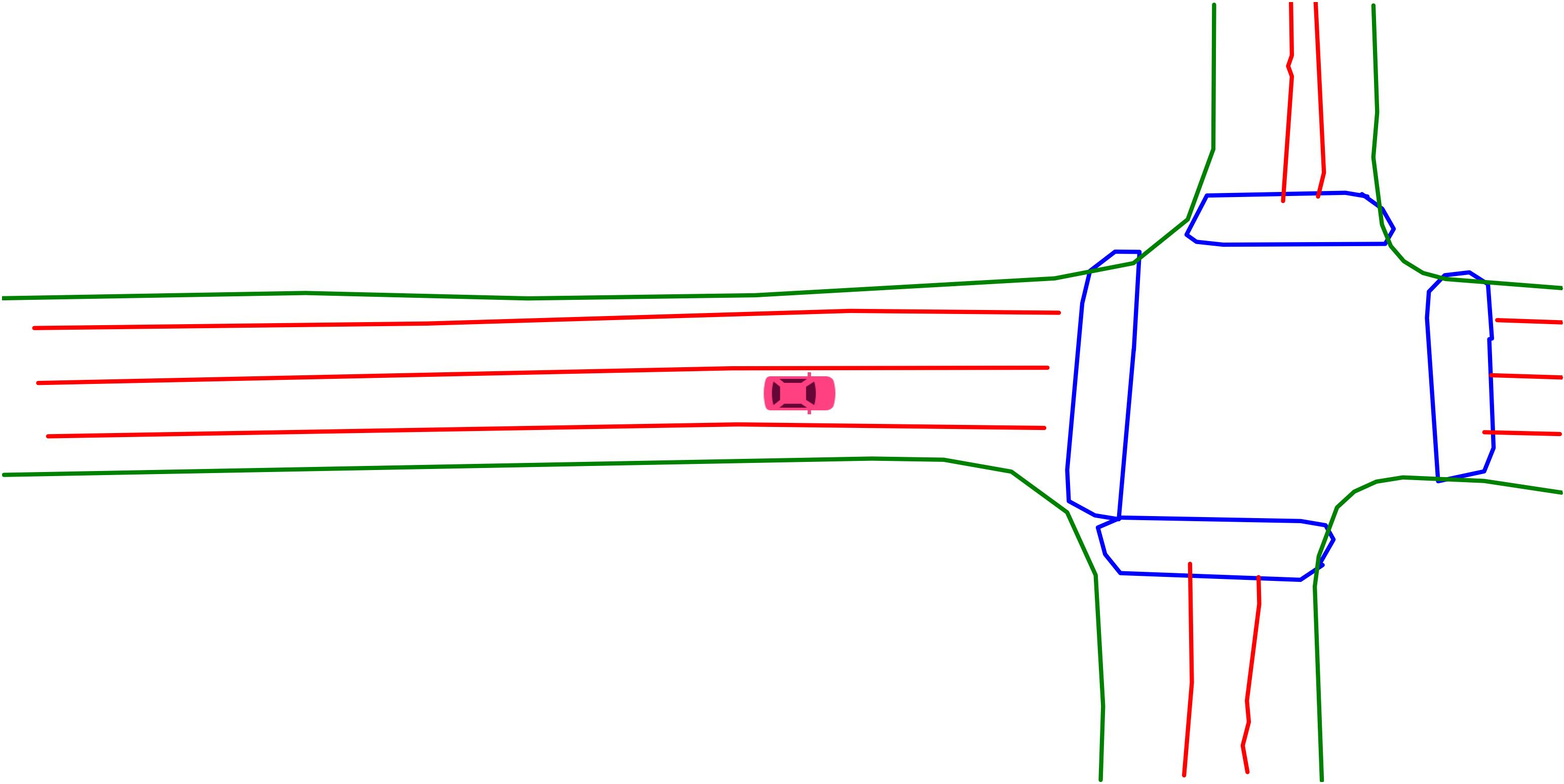}
    \end{subfigure} \\

    \begin{subfigure}{0.42\linewidth}
      \centering
      \includegraphics[width=\linewidth]{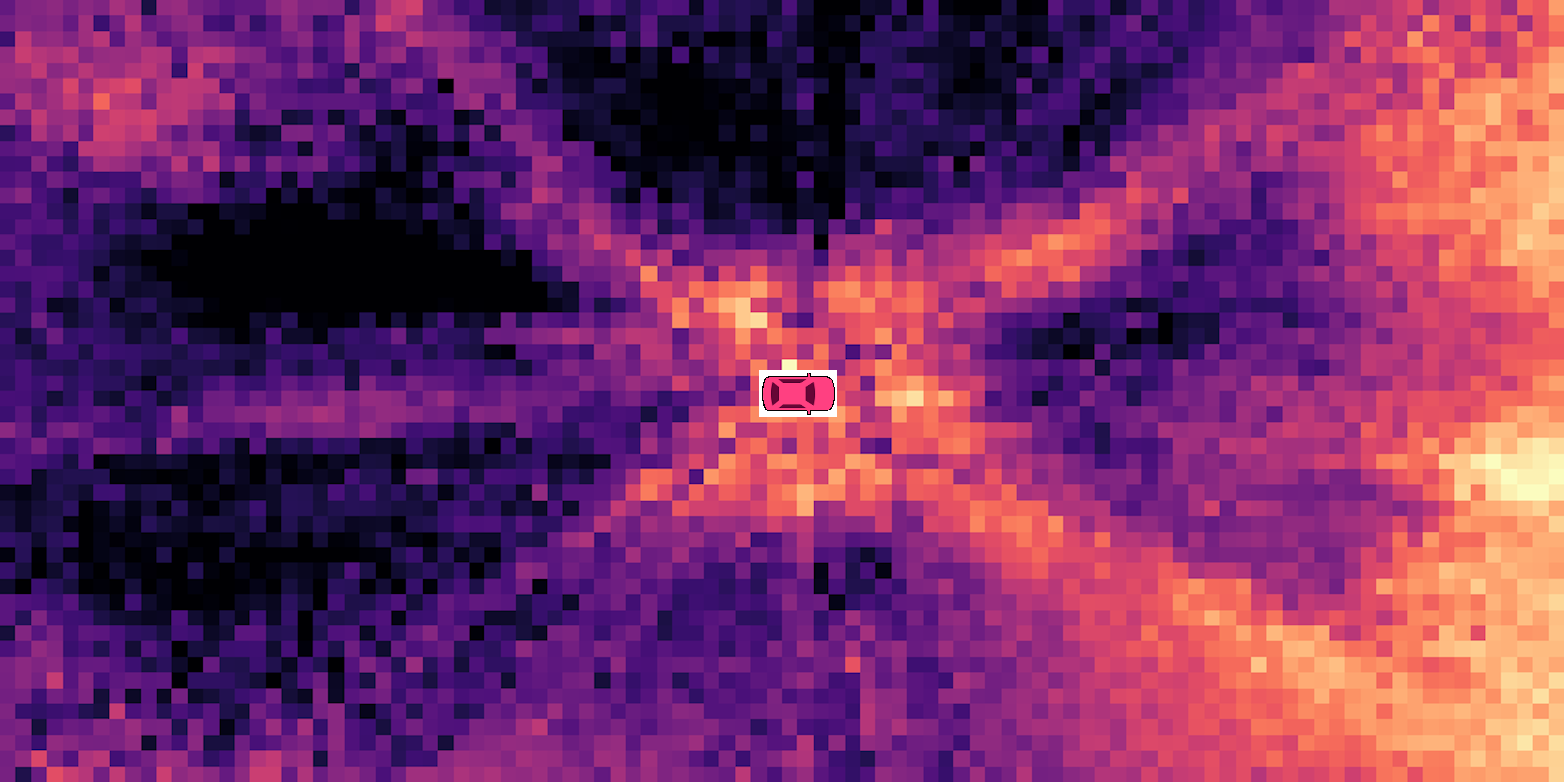}
    \end{subfigure} &
    \begin{subfigure}{0.42\linewidth}
      \centering
      \includegraphics[width=\linewidth]{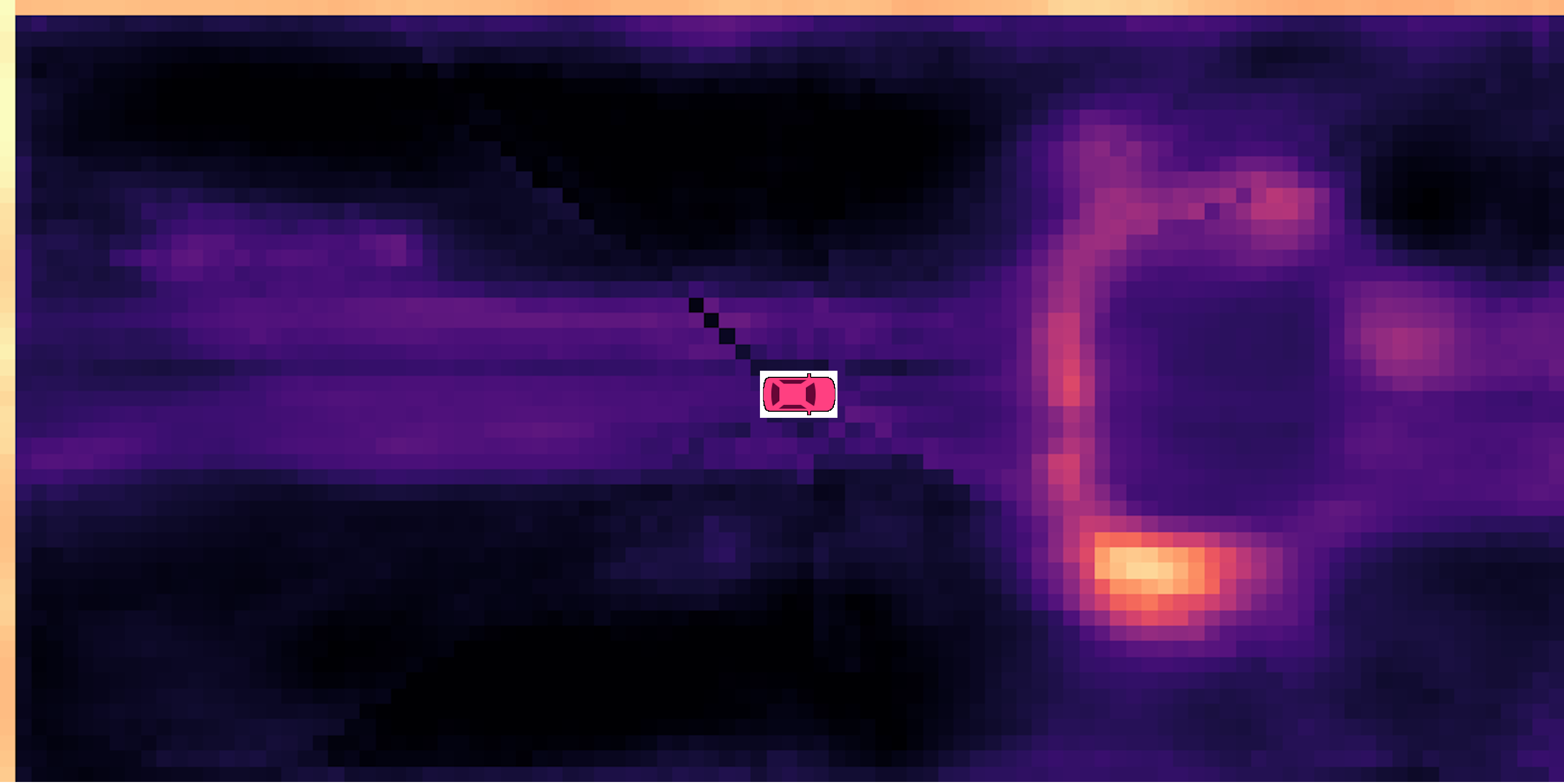}
    \end{subfigure}
  \end{tabular}

\caption{
    Qualitative comparison of map predictions and corresponding \gls{bev} feature activations.
    Top: predicted vector maps. Bottom: \gls{bev} feature visualizations.
    \gls{bev} visualizations show normalized channel-wise mean activations with shared scaling.
    }
  \label{fig:aerial_bev_features}
\end{figure}

\gls{hd} maps have long served as a key enabler of autonomous driving, providing geometric and semantic priors for localization, motion forecasting, and behavior planning~\cite{mapingoverview, poggenhans2018lanelet2, SmartMOT, HDMap3}.
However, they are expensive to generate, labor-intensive to maintain, and difficult to scale in dynamic environments with temporary traffic changes, construction, or seasonal variation~\cite{hdmapchallenges, bao2022highdefinitionmapgenerationtechnologies}.

To overcome these constraints, research has increasingly shifted toward mapless or map-light perception systems that infer structural priors directly from onboard sensor data~\cite{End2EndChalFront}.
A particularly promising direction lies in online map construction, where autonomous vehicles dynamically predict local maps from their current sensor observations, reducing dependence on pre-built maps while preserving spatial reasoning capabilities.

Recent advances in \acrfull{bev} perception have enabled camera-based systems to reconstruct road layouts and scene topology in real time from multi-camera observations.
Modern approaches~\cite{Li2021HDMapNet, Liao2023MapTR, Yuan2024StreamMapNet} convert multi-camera input into structured, vectorized outputs and demonstrate strong performance on autonomous driving benchmarks.
Nonetheless, maintaining geometric continuity and topological consistency at larger spatial extents or in complex scenes with occlusions, limited context, or sparsely visible cues remains challenging for camera-based \gls{bev} mapping approaches that rely on ego-centric observations~\cite{Yuan2024StreamMapNet}.

These challenges stem from the inherently ego-centric nature of camera-based perception.
The vehicle observes the scene from a limited field of view, so global spatial structure must be inferred from partial, perspective-distorted evidence.
Consequently, \gls{bev} encoders may produce locally inconsistent or fragmented representations, especially in distant regions where visual cues become sparse and information density decreases.
Moreover, supervision typically acts only through downstream task losses after decoding into semantic map outputs, providing indirect guidance that does not explicitly constrain the intermediate \gls{bev} feature geometry~\cite{Ye2025BEVDiffuser, Le2024DifFUSER, DiffBEV}.
This raises the question of how \gls{bev} encoders can be guided toward globally consistent spatial representations during training.

Aerial imagery provides a complementary overhead perspective in which road layout, connectivity, and long-range spatial relationships are directly observable.
Recent dataset extensions~\cite{Lengerer_AID4AD_2025} align aerial imagery with benchmarks such as nuScenes~\cite{nuScenes}, establishing pixel-level correspondence between aerial and ego-centric \gls{bev} representations.
These cross-view annotations enable aerial imagery to be used not only in fusion-based inference settings, but also as a training signal for supervising \gls{bev} representations across viewpoints.

In this work, we propose a training-time, feature-level \gls{cvs} method that transfers geometric and topological structure from aerial imagery to camera-based \gls{bev} encoders. Importantly, this supervision operates purely at the representation level during training and does not introduce additional inputs, modules, or constraints at inference time.
A pretrained aerial encoder provides dense \gls{bev} features from overhead views, which are used as a frozen teacher to guide the camera encoder through a lightweight alignment loss.
This training strategy allows the model to internalize structural context that would otherwise be missing from ego-centric observations, leading to sharper, more coherent \gls{bev} representations that improve map continuity and global layout, particularly in distant regions where sensor evidence becomes sparse.

Figure~\ref{fig:aerial_bev_features} illustrates these improvements, showing that relevant structural elements are already more prominent at the feature level, enabling better map predictions across the entire scene.
Since aerial supervision is used only during training, the final model remains unchanged in architecture, runtime, or input requirements.
This makes the trained model compatible with existing \gls{bev} pipelines without additional inference-time sensors, infrastructure, or computational overhead.

Our contributions are summarized as follows:
\begin{itemize}

\item We propose a \acrlong{cvs} strategy that transfers geometric and topological structure from a perspective-privileged aerial view to camera-based \gls{bev} encoders via dense feature alignment in \gls{bev} space.

\item We demonstrate that ego-aligned aerial imagery enables dense feature-level supervision for \gls{bev} encoders, providing a learnable alternative to conventional auxiliary losses based on semantic map projections.

\item Our method operates purely at training time, leaving the inference architecture, runtime, and input requirements unchanged.

\item We validate the approach on nuScenes using ego-aligned aerial imagery from the AID4AD cross-view extension, improving structural accuracy and spatial coherence over StreamMapNet without additional inference cost.

\end{itemize}

\section{Related Work}
\label{sec:relatedwork}

\subsection{Online HD Map Construction}

\gls{hd} maps provide geometric priors essential for high-level perception, yet their manual creation is costly and time-consuming~\cite{hdmapchallenges}. 
This motivates learning-based online mapping directly from onboard sensors. 
Early \gls{bev} approaches such as HDMapNet~\cite{Li2021HDMapNet}, Lift-Splat-Shoot~\cite{Philion2020LiftSplat}, and Pyramid Occupancy Networks~\cite{Roddick_2020_CVPR} represent the scene as rasterized occupancy grids. 
More recent vectorized methods, including VectorMapNet~\cite{Liu2023VectorMapNet}, MapTR~\cite{Liao2023MapTR}, and MapTRv2~\cite{MapTRv2}, improve precision and compactness by predicting structured map elements such as lane boundaries or dividers as polylines or keypoints. 
Subsequent works further enhance geometric consistency through curve-based parameterizations and instance-level structural constraints~\cite{Qiao_2023_CVPR, Liu_2024_CVPR}.

\subsection{Temporal Modeling in BEV Perception}

Transformer-based \gls{bev} architectures form the backbone of modern multi-camera perception. 
BEVFormer~\cite{Li2022BEVFormer} aggregates multi-view image features into a unified \gls{bev} representation via spatial cross-attention, while subsequent works extend this idea with temporal modeling. 
BEVDet4D~\cite{huang2022bevdet4d} stacks features across frames, and StreamPETR~\cite{Wang2023StreamPETR} propagates latent queries recurrently to maintain temporal context. 
StreamMapNet~\cite{Yuan2024StreamMapNet} adapts this paradigm to online \gls{hd} map construction by maintaining a streaming \gls{bev} state. 
Built upon the widely adopted BEVFormer backbone, it has become a common baseline for camera-based online map construction. 
Subsequent works further strengthen temporal consistency through persistent map representations, as explored by MapTracker~\cite{MapTracker} and Mask2Map~\cite{Mask2Map}.

\subsection{Cross-Modal Supervision and Diffusion-Based Structural Priors}

Recent work improves \gls{bev} perception through auxiliary supervision and structural regularization.  
Distillation-based approaches such as DistillBEV~\cite{Wang2023DistillBEV} and BEV-LGKD~\cite{Li2022BEVLGKD} transfer LiDAR-based geometric cues into camera encoders to improve spatial reasoning while maintaining vision-only inference.  
MapDistill~\cite{Hao2024MapDistill} extends this idea to online HD map construction by distilling fused LiDAR–camera features into a camera-only model, while SQD-MapNet~\cite{Wang2024SQDMapNet} stabilizes vectorized decoding through denoising-based refinement of noisy query predictions.

Diffusion-based approaches introduce iterative generative refinement for structural regularization.  
BEVDiffuser~\cite{Ye2025BEVDiffuser} formulates denoising as a layout-to-BEV generation task, while DifFUSER~\cite{Le2024DifFUSER} performs test-time refinement of fused features under sparse sensing.  
MapDiffusion~\cite{Monninger2025MapDiffusion} applies diffusion during decoding to improve robustness and uncertainty estimation.

One-for-All~\cite{OneForAll} systematically analyzes homogeneous and heterogeneous knowledge distillation setups and shows that mismatches in feature statistics and inductive biases can hinder direct feature transfer. 
This observation motivates lightweight alignment mechanisms when transferring representations across heterogeneous views.

\subsection{Cross-View Supervision with Aerial Imagery}

Aerial imagery has been used to extract road topology and map elements directly from overhead views, e.g., RoadTracer~\cite{roadtracer} and Sat2Graph~\cite{sat2graph}, and for ground-to-aerial matching in cross-view localization, e.g., CVM-Net~\cite{cvmnet}. 
Unlike these works, CVS neither constructs maps from aerial imagery nor performs cross-view retrieval; instead, it uses ego-aligned aerial features as train-time supervision for camera-only online HD map construction.

Aerial imagery provides a complementary top-down perspective that captures global road layout, long-range connectivity, and large-scale spatial context. 
The AID4AD dataset~\cite{Lengerer_AID4AD_2025} enables pixel-level cross-view supervision by aligning aerial imagery with the nuScenes ego coordinate frame. 
Fusion-based experiments show that overhead context improves structural completeness and large-scale geometric consistency, but require dual encoders and continuous aerial availability.

In contrast, CVS uses aerial imagery only during training to supervise the camera-based \gls{bev} representation through feature-level alignment, avoiding additional sensors or inputs at inference time.

\section{Methodology}
\label{sec:aerialframework}

\begin{figure*}[t]
    \centering
    \includegraphics[width=0.98\textwidth]{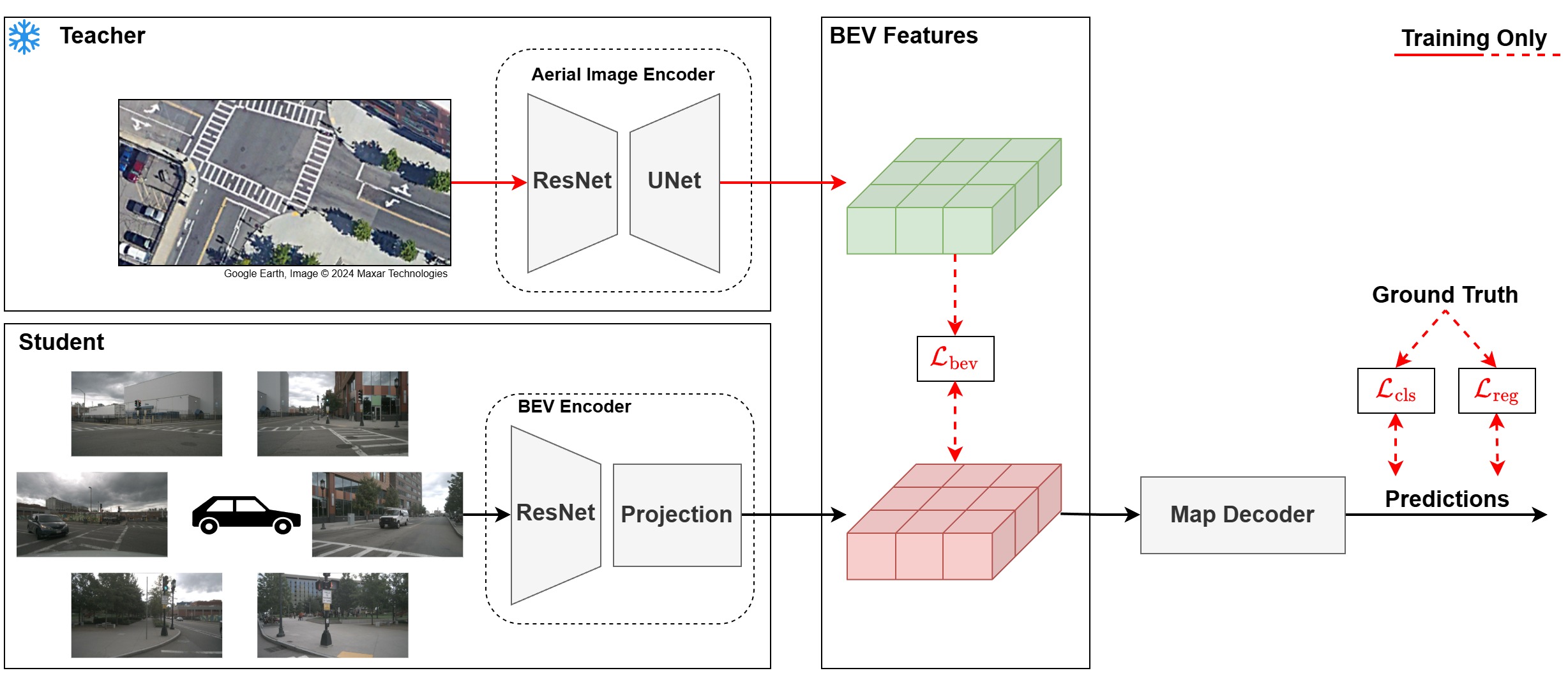}
    \caption{Overview of the proposed aerial-guided \gls{bev} training approach. 
    During training, aerial imagery provides feature-level guidance for the camera-based encoder through the \gls{bev} loss $\mathcal{L}_{\text{bev}}$. 
    At inference, only the camera stream is active, identical in structure to StreamMapNet.}
    \label{fig:meth_overview}
\end{figure*}

We introduce a training framework that transfers geometric and topological priors from overhead imagery into camera-based \gls{bev} encoders through cross-view supervision.  
Unlike sensor-fusion approaches that require aerial inputs during inference, our method uses aerial imagery only as a training signal, preserving real-time camera-only operation at deployment.

We instantiate this idea within the StreamMapNet framework~\cite{Yuan2024StreamMapNet}, a strong baseline for online \gls{hd} map construction.  
As illustrated in \Cref{fig:meth_overview}, our approach augments the camera-based mapping pipeline with an auxiliary aerial supervision branch that provides structured guidance during training while leaving the inference architecture unchanged.

\subsection{Overview}
\label{subsec:overview}

The aerial encoder \(E^{\text{aerial}}\) serves as a frozen teacher and produces structured \gls{bev} features \(F^{\text{aerial}} \in \mathbb{R}^{C \times H \times W}\) from geo-referenced overhead imagery. 
The camera encoder \(E^{\text{cam}}\) serves as the student, generating \(F^{\text{cam}} \in \mathbb{R}^{C \times H \times W}\) from multi-camera input, which is aligned with \(F^{\text{aerial}}\) during training via a lightweight MSE loss.

Both feature maps share the same spatial shape \(C \times H \times W = 256 \times 50 \times 100\), where \(C\) denotes the channel dimension and \(H \times W\) the spatial resolution of the \gls{roi}.
The AID4AD alignment pipeline provides accurate pixel-level correspondence between aerial and ego-centric \gls{bev} representations, enabling precise cross-view supervision.

\subsection{Cross-View BEV Supervision}

We align the \gls{bev} representations of both encoders during training through feature-level cross-view supervision.
A pretrained aerial encoder serves as a frozen teacher that provides structurally consistent \gls{bev} features, while the camera-based encoder learns to match these representations through a feature alignment objective.
This formulation enables direct supervision in the dense \gls{bev} feature space by leveraging perspective-aligned aerial imagery as a structured teacher signal.

The aerial encoder \(E^{\text{aerial}}\) encodes ego-aligned aerial crops \(I^{\text{aerial}}\) into dense \gls{bev} features \(F^{\text{aerial}}\):

\begin{equation}
    F^{\text{aerial}} = E^{\text{aerial}}(I^{\text{aerial}})
\end{equation}

We employ an aerial encoder trained on the AID4AD dataset~\cite{Lengerer_AID4AD_2025} for \gls{bev}-based map construction from overhead imagery.
The network follows a ResUNet architecture with a ResNet backbone~\cite{resnet} and U-Net decoder~\cite{unet}, producing metrically consistent \gls{bev} representations from high-resolution aerial imagery.

These features capture lane boundaries, road edges, and intersection topology, providing geometric cues that serve as structural supervision for the camera-based \gls{bev} encoder. 
The aerial perspective offers a geometrically consistent view of the scene, enabling the transfer of global spatial structure that is difficult to infer from ego-centric observations alone.

The camera-based encoder \(E^{\text{cam}}\) adopts the BEVFormer-style backbone of StreamMapNet. 
Given \(N{=}6\) synchronized camera views \(I_{1:N}\), it encodes multi-view features into a unified \gls{bev} representation \(F^{\text{cam}}\) via deformable attention:
\begin{equation}
    F^{\text{cam}} = E^{\text{cam}}(I_{1:N})
\end{equation}

To address scale and shift discrepancies between aerial and camera feature distributions, 
we introduce an Affine Adapter in the student branch that applies a per-channel affine transformation to the camera feature tensor used for supervision.  
Let \(F^{\text{cam}} \in \mathbb{R}^{C \times H \times W}\) denote the \gls{bev} feature map from the camera encoder.  
The adapter transforms each channel independently according to
\begin{equation}
\tilde{F}^{\text{cam}}_{i,:,:} = \gamma_i \cdot F^{\text{cam}}_{i,:,:} + \beta_i, \quad \text{for } i = 1, \dots, C
\end{equation}
where \(\gamma, \beta \in \mathbb{R}^C\) are learnable parameters representing channel-wise scale and shift.  
The adapter is active only within the loss path during training and leaves the inference-time encoder unchanged.

Prior to computing the alignment loss, both \gls{bev} feature maps are channel-wise normalized to mitigate scale bias and stabilize optimization. 
The resulting \gls{bev} alignment loss is implemented as a \gls{mse} objective:

\begin{equation}
\mathcal{L}_{\text{bev}} =
\frac{1}{CHW}
\sum_{c=1}^{C}
\sum_{h=1}^{H}
\sum_{w=1}^{W}
\left(
\tilde{F}^{\text{cam}}_{c,h,w}
-
F^{\text{aerial}}_{c,h,w}
\right)^2
\end{equation}

The adapted representation $\tilde{F}^{\text{cam}}$ is aligned with the aerial \gls{bev} features through the supervision loss, encouraging the student encoder to internalize the spatial structure and continuity encoded in the aerial representation.

\subsection{Training Objective}
Following StreamMapNet, the decoder predicts vectorized map elements using a classification head optimized with Focal Loss~\cite{Lin2017FocalLoss} (\(\mathcal{L}_{\text{cls}}\)) and a geometry head trained with a line-based L1 loss (\(\mathcal{L}_{\text{reg}}\)).  

The overall training objective extends this baseline with an auxiliary \gls{bev}-guided loss:
\begin{equation}
    \mathcal{L}_{\text{total}} =
    \mathcal{L}_{\text{cls}} +
    \mathcal{L}_{\text{reg}} +
    \lambda_{\text{bev}} \mathcal{L}_{\text{bev}}
\end{equation}
where \(\lambda_{\text{bev}}\) controls the relative strength of aerial supervision that aligns the camera encoder’s feature representation with the aerial domain, balancing feature-level guidance with the primary map prediction objectives.

\subsection{Dataset and Evaluation}
AID4AD~\cite{Lengerer_AID4AD_2025} provides aerial imagery registered to the nuScenes dataset local coordinate frame, 
ensuring pixel-level correspondence between aerial imagery and ego-centric \gls{bev} representations.
The task follows the standard online \gls{hd} map construction setting, 
where the model predicts structured, vectorized map elements within a defined \gls{roi}, including road boundaries, lane dividers, and crosswalks.

We follow StreamMapNet~\cite{Yuan2024StreamMapNet} and adopt the geographically separated data split introduced by Roddick and Cipolla~\cite{Roddick_2020_CVPR}. 
StreamMapNet further quantified that the original nuScenes split exhibits a high geographic overlap between train and test frames (approximately 84\%), 
highlighting the importance of evaluating generalization under geographically separated conditions.

Performance is reported as \gls{map} for each semantic class and for the overall score. 
The semantic classes include pedestrian crossings (AP$_{\text{ped}}$), road dividers (AP$_{\text{div}}$) and lane boundaries (AP$_{\text{bound}}$).
Following prior works~\cite{Li2021HDMapNet, Liu2023VectorMapNet, Yuan2024StreamMapNet}, we evaluate two regions of interest: 
\(60\times30~\mathrm{m}\), matching most \gls{bev}-based approaches, and a larger \(100\times50~\mathrm{m}\) region to assess long-range generalization.
The \gls{map} is computed as the average over distinct distance thresholds, 
using \(\{0.5, 1.0, 1.5\}\,\mathrm{m}\) for the \(60\times30~\mathrm{m}\) region 
and \(\{1.0, 1.5, 2.0\}\,\mathrm{m}\) for the \(100\times50~\mathrm{m}\) region.

\subsection{Implementation Details}

We use AdamW with an initial learning rate of \(1.25\times10^{-4}\), cosine annealing, and a batch size of 4 on a single NVIDIA L40S GPU.  
The learning rate is adjusted for single-GPU training, while all other architectural, training, and evaluation settings remain identical to StreamMapNet, ensuring that observed differences can be attributed solely to aerial-guided supervision.  
Unless otherwise stated, the supervision weight is set to \(\lambda_{\text{bev}} = 60\) for the \(60\times30\,\mathrm{m}\) \gls{roi} and \(\lambda_{\text{bev}} = 70\) for the extended \(100\times50\,\mathrm{m}\) setting.
 
\subsection{Feature Space Analysis}
\label{subsec:feature_similarity}

While improved mAP confirms the effectiveness of aerial supervision, it does not reveal how the internal representations of the \gls{bev} encoder change.
Since our approach supervises feature representations rather than semantic predictions, we analyze how aerial guidance influences the structure of the learned \gls{bev} feature space.
To understand the role of normalization and affine adaptation in stabilizing cross-view feature alignment, we evaluate teacher–student feature similarity across four training variants:
\begin{enumerate}
    \item baseline StreamMapNet (no supervision)
    \item supervision without normalization or affine adaptation
    \item supervision with normalization only
    \item full cross-view supervision setup with normalization and affine adaptation
\end{enumerate}
All metrics are computed on the full validation set to ensure statistically robust estimates.

\medskip
\noindent\textbf{Similarity Metrics.}\, Representation similarity is commonly analyzed using canonical correlation methods
(SVCCA~\cite{raghu2017svcca}, PWCCA~\cite{morcos2018pwcca}) 
and linear regression–based measures.  
Kornblith \textit{et al.}~\cite{kornblith2019similarity} systematically compare these approaches 
and introduce linear \gls{cka} as a robust and interpretable metric for comparing neural activations across architectures.  
\gls{cka} is invariant to isotropic scaling and orthogonal transformations 
and captures structural rather than pointwise similarity between feature spaces.

Following~\cite{kornblith2019similarity,OneForAll}, we compute linear \gls{cka} between 
the aerial teacher features and each student \gls{bev} feature variant.  
Let \(X\) denote the teacher feature matrix and \(Y\) the corresponding student feature matrix;  
then
\begin{equation}
\mathrm{CKA}(X,Y) = 
\frac{\| X^\top Y \|_F^2}{
\| X^\top X \|_F \, \| Y^\top Y \|_F},
\qquad
X,Y \in \mathbb{R}^{N\times C}
\label{eq:cka}
\end{equation}
Higher \gls{cka} values indicate that the student encoder organizes information 
in a feature space more structurally aligned with the aerial teacher.

\medskip
\noindent\textbf{Coefficient of Determination (R\textsuperscript{2}).}\,
In addition to \gls{cka}, we compute the coefficient of determination ($R^2$) 
to assess how well teacher features can be linearly reconstructed from student features.

This regression-based measure complements \gls{cka} by quantifying the strength of direct value correspondence, which closely relates to the \gls{mse} used as training loss. 
Together, these metrics allow us to distinguish between direct feature imitation and structural alignment in the latent space, providing insight into how aerial supervision shapes the \gls{bev} representation beyond improvements in task performance.

\section{Results}
\label{sec:experiments}

We evaluate whether cross-view supervision improves camera-based BEV mapping while preserving the camera-only inference pipeline.
The following experiments analyze both quantitative performance and the resulting changes in the learned \gls{bev} representations.

\subsection{Quantitative Results}
\begin{table*}[t]
\centering
\caption{
Per-class and overall AP (\%) on AID4AD under identical camera-only inference.
Relative improvements (shown next to mAP) highlight the particularly strong gains in the extended \(100\times50\,\mathrm{m}\) region.
}
\renewcommand{\arraystretch}{1.05}
\setlength{\tabcolsep}{4pt}
\begin{tabular}{l l c c c l}
\toprule
RoI & Method &
AP$_{\text{ped}}$$\uparrow$ &
AP$_{\text{div}}$$\uparrow$ &
AP$_{\text{bound}}$$\uparrow$ &
mAP$\uparrow$ \\
\midrule
 & StreamMapNet~\cite{Yuan2024StreamMapNet} &
32.2 & 29.3 & 40.8 &
34.1\;{\color{DarkGray}(--)} \\

60$\times$30 & \textbf{StreamMapNet+CVS} &
40.1 & 30.3 & 43.5 &
38.0\;{\color{DarkGreen}\textbf{(+11\%)}} \\

\cmidrule(lr){2-6}
 & \textit{$\Delta$ (CVS $-$ baseline)} &
+7.9 & +1.0 & +2.7 & +3.9 \\
\midrule
 & StreamMapNet~\cite{Yuan2024StreamMapNet} &
25.6 & 17.4 & 24.3 &
22.4\;{\color{DarkGray}(--)} \\

100$\times$50 & \textbf{StreamMapNet+CVS} &
40.3 & 25.8 & 30.7 &
32.3\;{\color{DarkGreen}\textbf{(+44\%)}} \\

\cmidrule(lr){2-6}
 & \textit{$\Delta$ (CVS $-$ baseline)} &
+14.7 & +8.4 & +6.4 & +9.9 \\
\bottomrule
\label{tab:main_results}
\end{tabular}
\end{table*}

\Cref{tab:main_results} summarizes the quantitative results on the AID4AD dataset under identical camera-only inference conditions.
Across both regions of interest, cross-view supervision consistently improves mapping accuracy over the StreamMapNet baseline.
Notably, the improvement becomes substantially larger at extended spatial ranges.
While gains are moderate within the standard \(60\times30\,\mathrm{m}\) region, the performance increase grows significantly in the \(100\times50\,\mathrm{m}\) setting, indicating that aerial supervision is particularly beneficial where ego-centric observations become sparse.

For the \(60\times30~\mathrm{m}\) region, performance increases from 34.1 to 38.0~mAP (+11.4\%), with consistent improvements across all semantic categories, indicating enhanced structural coherence of the learned \gls{bev} representations.

For the larger \(100\times50~\mathrm{m}\) region, the benefit becomes even more pronounced: performance rises from 22.4 to 32.3~mAP (+9.9 absolute, +44.2\% relative), highlighting the effectiveness of aerial supervision in long-range scenarios where camera observations become sparse and structurally ambiguous.

To compare against widely used auxiliary supervision, we train StreamMapNet with a MapTRv2-style auxiliary loss on the same \(100\times50~\mathrm{m}\) setup, supervising the \gls{bev} feature through a lightweight segmentation head and GT-derived target. 
The auxiliary baseline yields only a limited improvement from 22.4 to 23.3~mAP; larger loss weights of 3 and 5 result in 22.4 and 22.8~mAP, respectively, remaining below CVS at 32.3~mAP.
This indicates that an auxiliary \gls{bev}-level objective alone does not account for the CVS gain.

These results suggest that aerial guidance improves the model’s ability to maintain global spatial consistency beyond the region directly supported by dense ego-sensor observations.
Overall, cross-view supervision transfers geometric context from aerial imagery into the camera-based \gls{bev} encoder, substantially improving large-scale mapping performance without modifying the inference pipeline.

To further analyze how aerial supervision affects the learned representations, we conduct controlled ablations that examine
(i) the role of normalization and the affine adapter,
(ii) the structural correspondence between aerial and camera feature spaces, and
(iii) the influence of the aerial teacher architecture.

\medskip
\noindent\textbf{Effect of Normalization and Affine Adapter}\, To isolate the effect of the proposed stabilization components, we train additional variants on the \(100\times50\,\mathrm{m}\) setting while keeping all other configurations fixed. 
Cross-view supervision without normalization or adaptation already increases performance to 28.9~mAP, indicating that aerial supervision provides useful structural guidance despite modality-induced feature shifts. 
Adding feature normalization yields a strong improvement to 31.5~mAP, highlighting that correcting scale discrepancies between modalities is essential for stable alignment. 
Introducing the affine adapter on top of normalization offers a further refinement, reaching 32.3~mAP. 
These results show that normalization resolves the dominant mismatch between camera and aerial features, while the affine adapter compensates residual channelwise offsets, leading to smoother optimization and more reliable cross-view supervision.

\begin{table}[t]
\centering
\caption{
Ablation study on feature normalization and the affine adapter in the student branch, evaluated on the extended \(100\times50\,\mathrm{m}\) region of interest.
All models use \(\lambda_{\text{bev}}{=}70\) and identical training settings.
Reported values denote overall mAP (\%) on the AID4AD dataset.
\(\Delta\) is computed relative to the StreamMapNet baseline (22.4 mAP).
}
\begin{tabular}{lcc}
\toprule
\textbf{Configuration} & \textbf{mAP}$\uparrow$ & $\Delta$ \\
\midrule
w/o normalization, w/o adapter & 28.9 & +6.5 \\
+ normalization only & 31.5 & +9.1 \\
+ normalization + affine adapter & \textbf{32.3} & \textbf{+9.9} \\
\bottomrule
\end{tabular}

\label{tab:adapter_ablation}
\end{table}

\begin{figure}[t]
    \centering
    \includegraphics[width=0.98\linewidth]{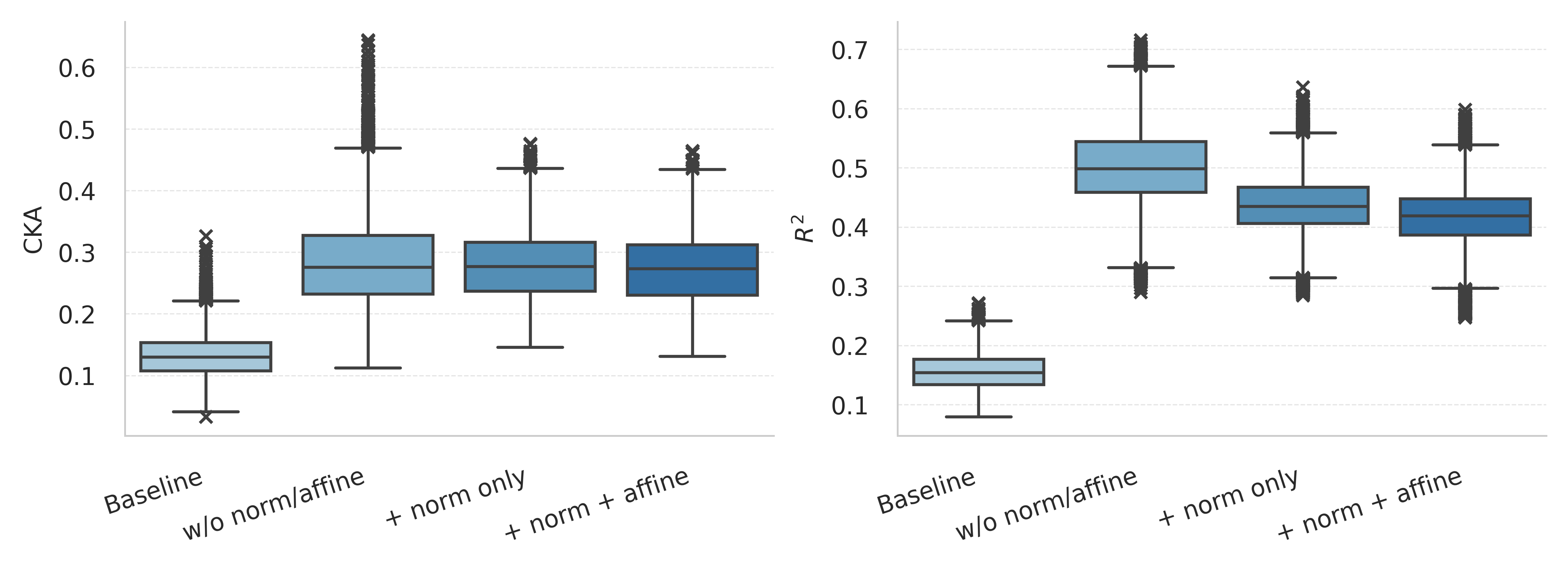}
    \caption{
    Distribution of feature similarity metrics (linear CKA and $R^{2}$) between teacher and student encoders. 
    Supervised variants share the same Cross-View Supervision setup and differ only in the projection module 
    (w/o normalization/affine, normalization only, normalization + affine adapter). 
    Higher values indicate stronger structural alignment.
    }
    \label{fig:feature_similarity}
\end{figure}

\noindent\textbf{Feature Similarity Analysis}\,
Figure~\ref{fig:feature_similarity} visualizes feature similarity between teacher and student encoders.  
All supervised variants outperform the baseline, confirming that aerial guidance improves cross-view feature alignment.  

Across supervised variants, median \gls{cka} remains comparable, while $R^2$ progressively decreases when normalization and affine adaptation are introduced.  
Although the \gls{bev} loss in Eq.~(4) is value-based in form, these components relax exact raw feature-value copying between heterogeneous teacher and student representations, consistent with prior findings on the importance of adaptation in heterogeneous distillation~\cite{OneForAll}.  
Together with the mAP ablation, this suggests that preserving spatially aligned activation structure, reflected by stable \gls{cka}, is more relevant than maximizing direct value reconstruction, reflected by $R^2$.

\medskip
\noindent\textbf{Influence of the Aerial Encoder}\,
To analyze how aerial teacher architecture affects cross-view supervision, we compare three teacher variants.
The additional ResUNet++ and UNet++ teachers are derived from the AID4AD fusion setup by removing the ego-view branch, yielding aerial-only teachers trained from aligned overhead imagery. 
This enables scenario-consistent yaw augmentation with six fixed rotations per scenario, increasing rotational diversity while preserving temporal coherence.

The original ResUNet teacher reaches 47.3~mAP in its teacher setup and yields a CVS student of 32.3~mAP.
Replacing it with ResUNet++ improves teacher-side performance to 52.2~mAP, but decreases the corresponding CVS student to 30.5~mAP.
By contrast, a UNet++ teacher reaches 55.0~mAP and yields a CVS student of 32.2~mAP, matching the original ResUNet-based CVS model.

These results suggest that standalone aerial mapping performance and cross-view transfer quality are related but distinct objectives.
Teacher-side accuracy alone is therefore not a reliable predictor of CVS student performance; the effectiveness of supervision depends on the transferability of the intermediate teacher representation to the camera-based \gls{bev} encoder.
Increasing the BEVFormer student encoder from 1 to 2 layers does not materially change the result, indicating that student capacity alone is not the limiting factor.
Together with the CKA/$R^2$ analysis in Fig.~\ref{fig:feature_similarity}, this suggests that successful cross-view supervision is governed by transferable spatial structure rather than teacher-side accuracy alone.
Appendix~\ref{app:teacher_features} provides qualitative teacher-feature visualizations supporting this interpretation.

\subsection{Qualitative Results}

\begin{figure*}[t]
\centering

\begin{minipage}{0.35\linewidth}
    \centering
    \includegraphics[width=\linewidth]{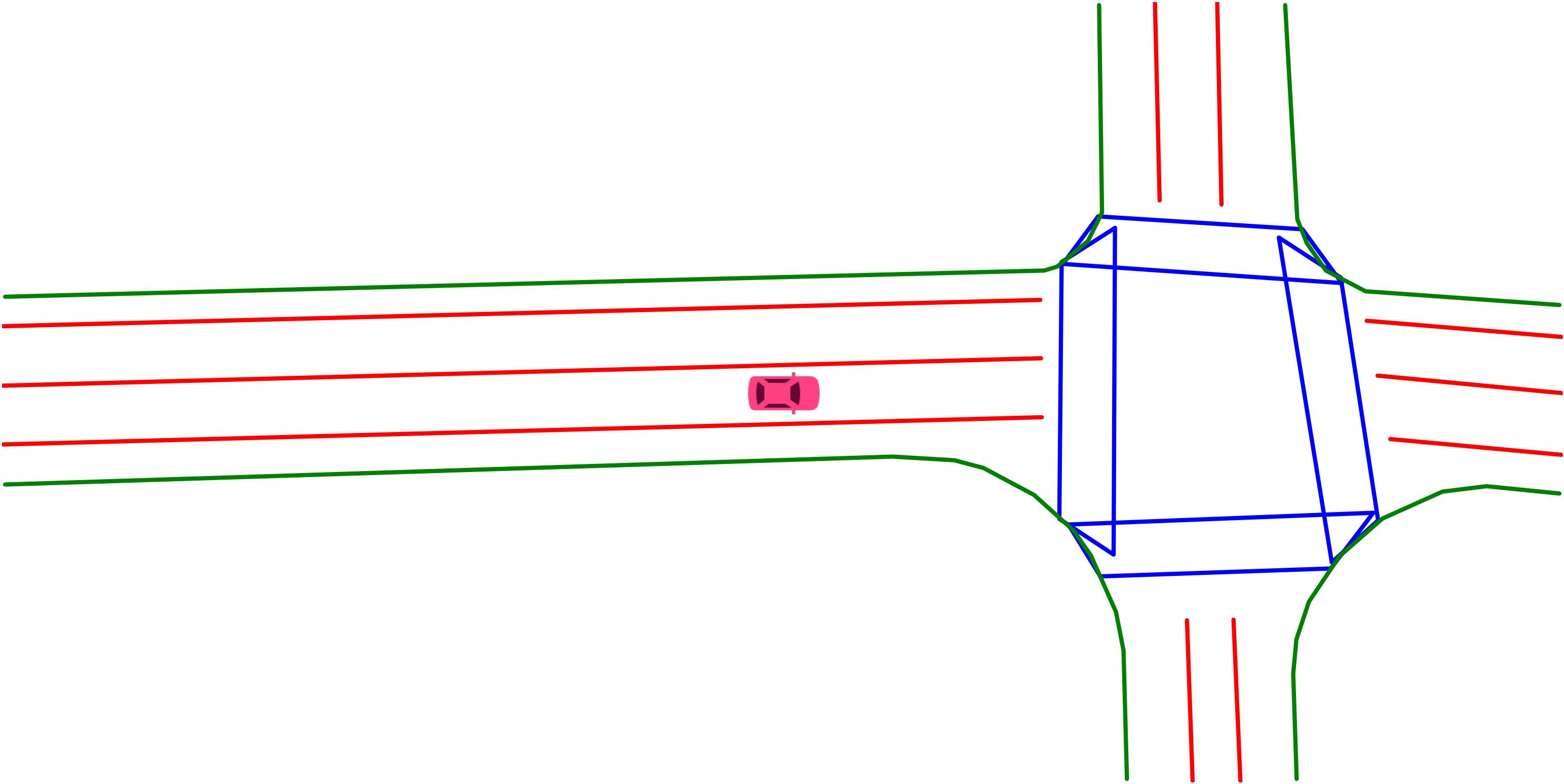}
    {\scriptsize Ground Truth}
\end{minipage}
\hspace{0.02\linewidth}
\begin{minipage}{0.35\linewidth}
    \centering
    \includegraphics[width=\linewidth]{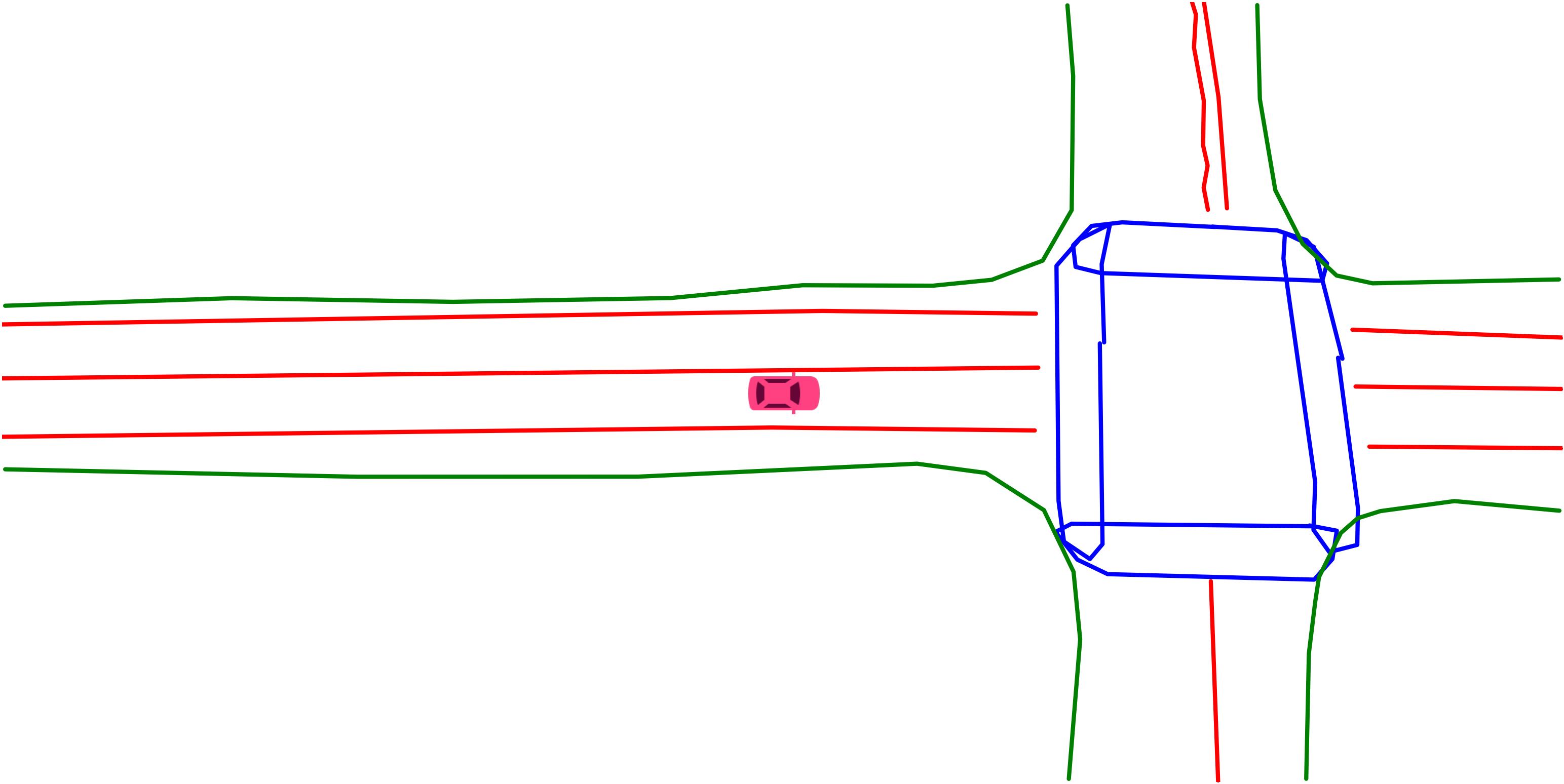}
    {\scriptsize Teacher (Fusion Model)}
\end{minipage}

\vspace{0.6em}

\begin{minipage}{0.35\linewidth}
    \centering
    \includegraphics[width=\linewidth]{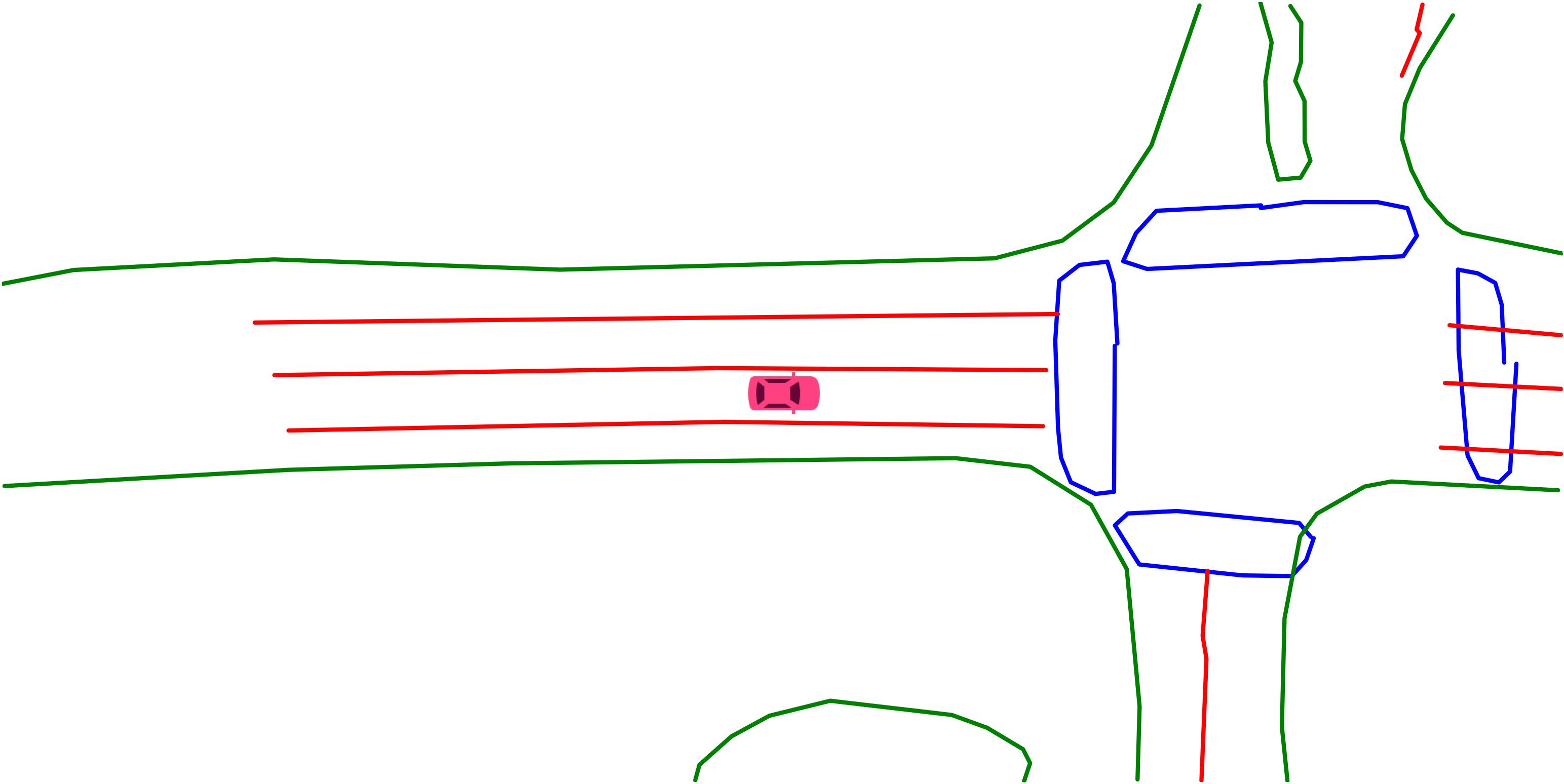}
    {\scriptsize Student w/o Aerial Guidance}
\end{minipage}
\hspace{0.02\linewidth}
\begin{minipage}{0.35\linewidth}
    \centering
    \includegraphics[width=\linewidth]{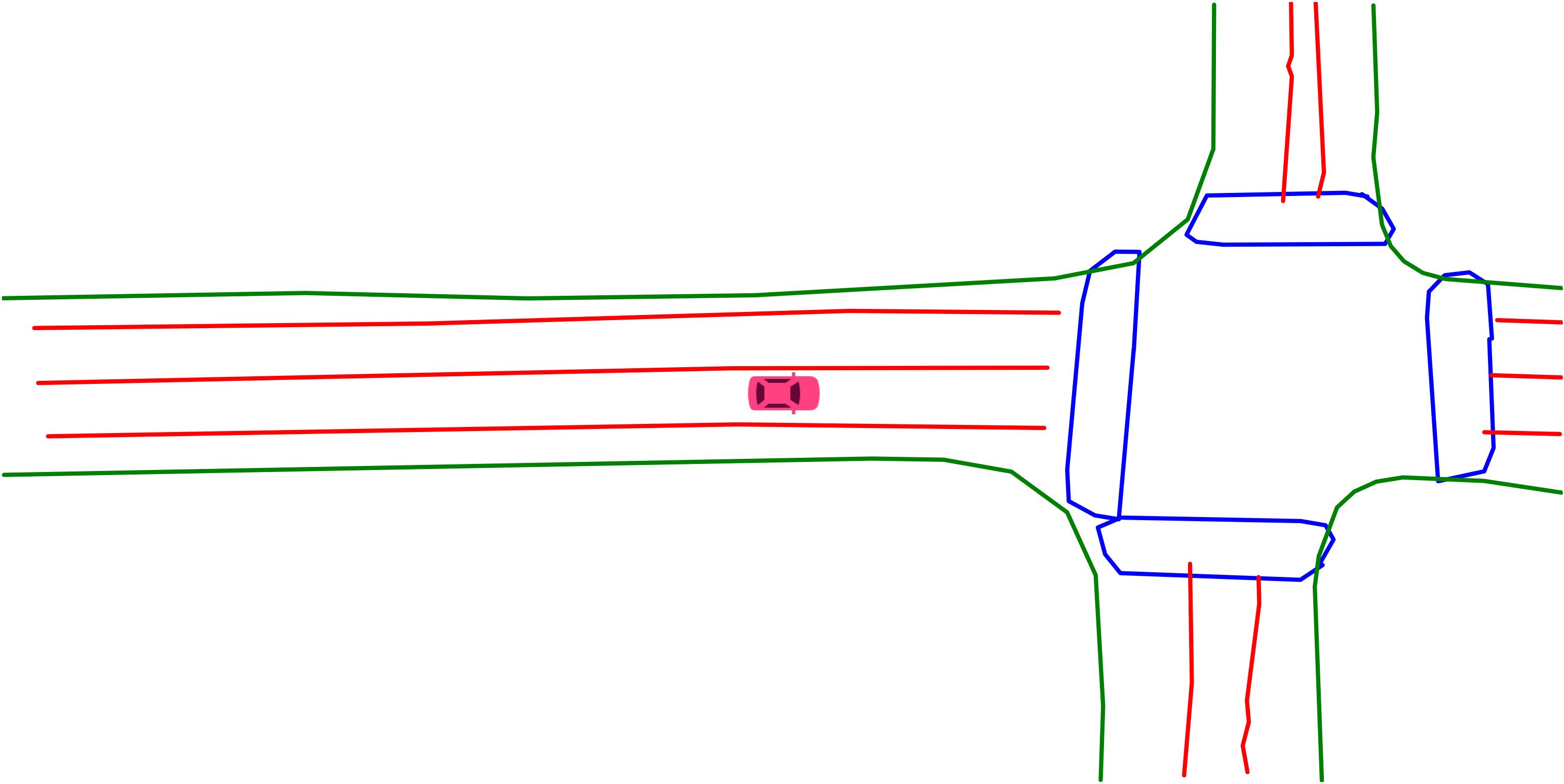}
    {\scriptsize Student with Aerial Guidance}
\end{minipage}

\caption{
Qualitative comparison of online HD map construction on nuScenes enriched with aerial imagery from AID4AD. Shown are ground truth, fusion teacher, baseline student, and aerial-guided student.
}
\label{fig:qualitative}

\end{figure*}

\begin{figure}[t]
  \centering
  \setlength{\tabcolsep}{2pt}
  \renewcommand{\arraystretch}{0}

  \begin{tabular}{cc}
    \includegraphics[width=0.33\linewidth]{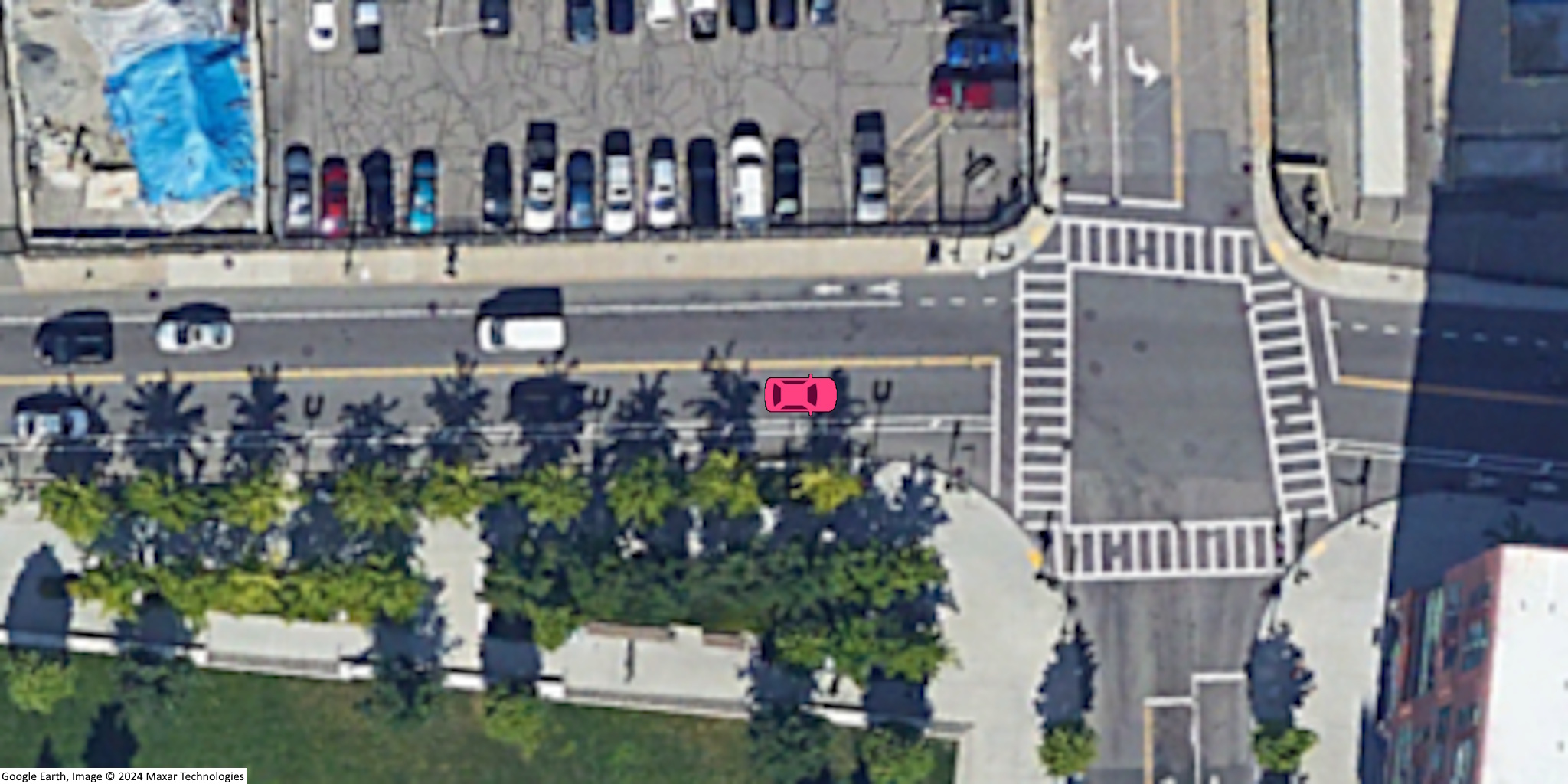} &
    \includegraphics[width=0.33\linewidth]{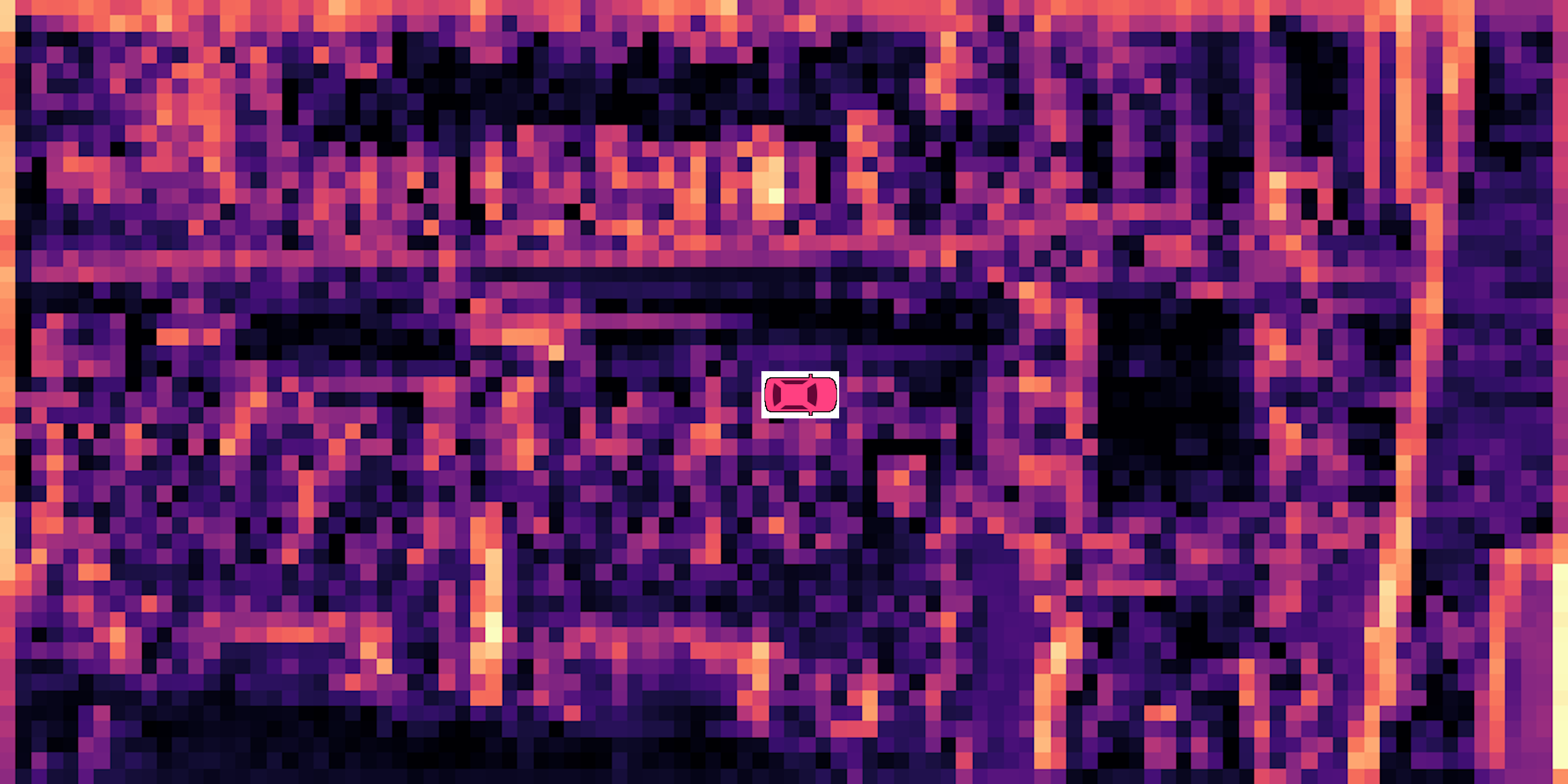} \\
    \parbox[t]{0.33\linewidth}{\centering \small \vspace{0.3em} Aerial RGB} &
    \parbox[t]{0.33\linewidth}{\centering \small \vspace{0.3em} Aerial BEV (teacher)} \\
    \\[0.8em]
    \includegraphics[width=0.33\linewidth]{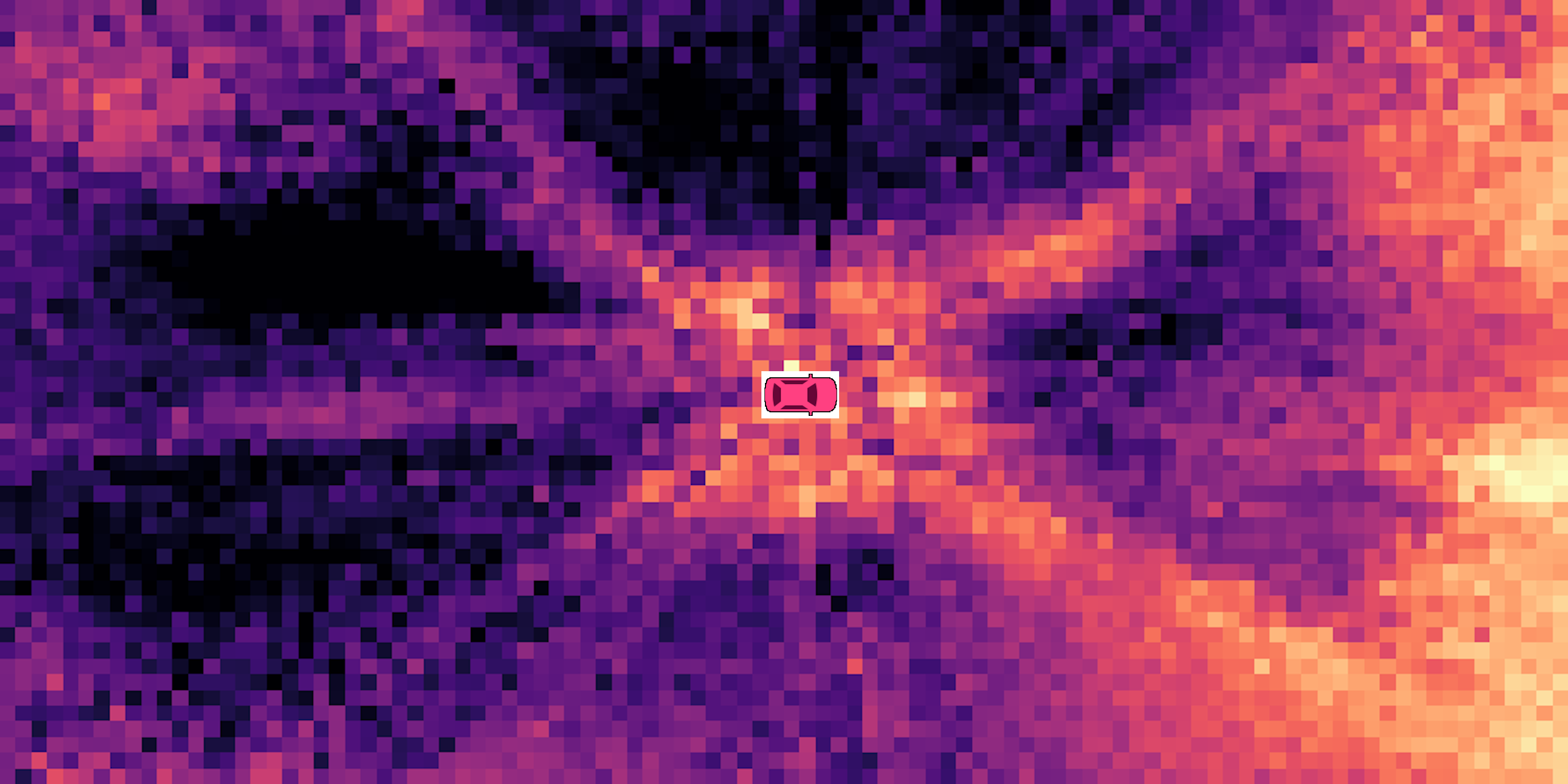} &
    \includegraphics[width=0.33\linewidth]{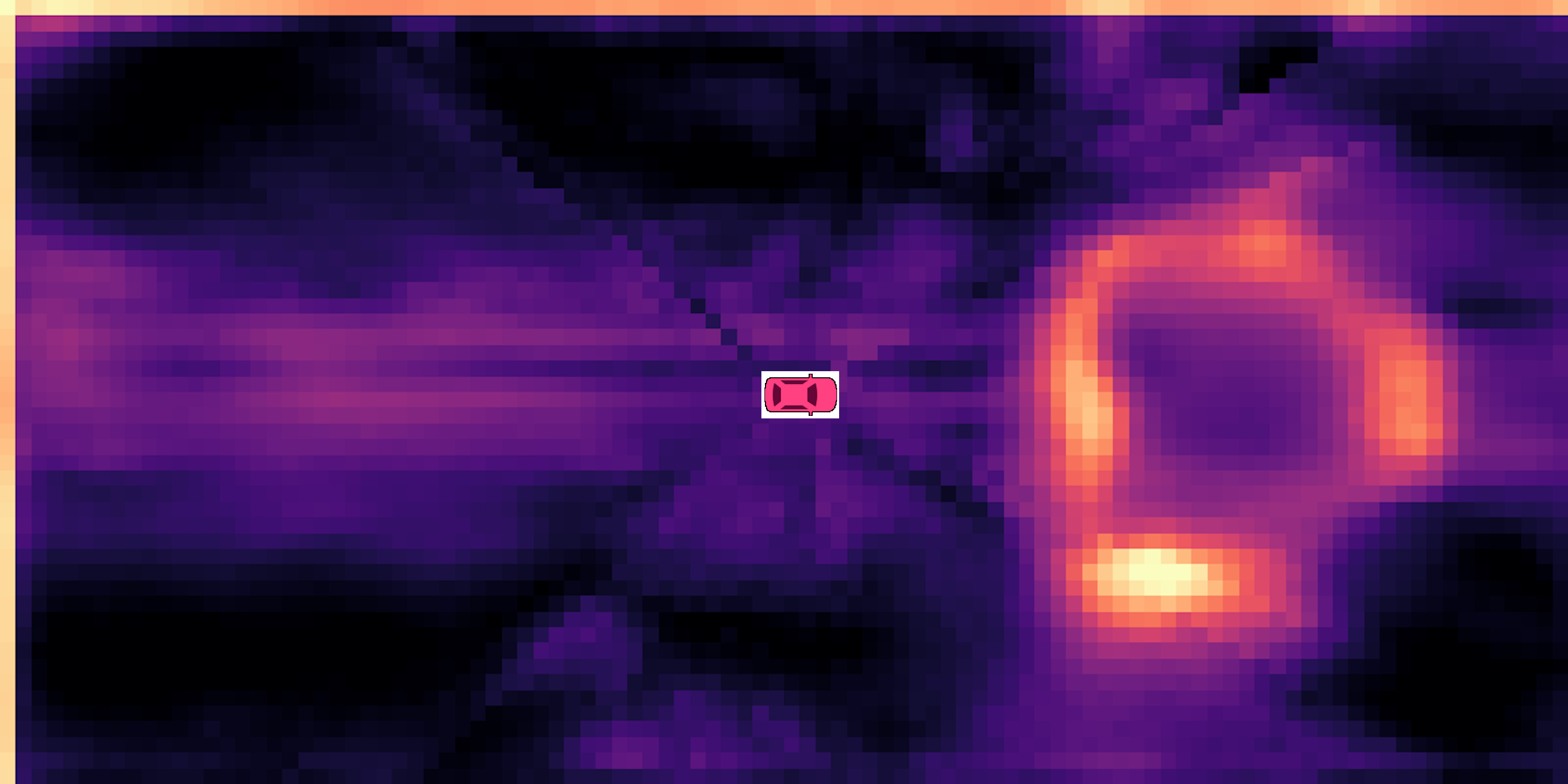} \\
    \parbox[t]{0.33\linewidth}{\centering \small \vspace{0.3em} Camera BEV (baseline)} &
    \parbox[t]{0.33\linewidth}{\centering \small \vspace{0.3em} Camera BEV (ours)}
  \end{tabular}

    \caption{
    Qualitative BEV feature visualization under cross-view supervision.
    Aligned aerial RGB imagery is encoded into structured teacher features (top row), which guide the camera encoder via feature-level BEV alignment during training.
    The guided camera features are sharper and more spatially coherent than the camera-only baseline (bottom row), especially under sparse or occluded visual evidence.
    }
  \label{fig:bev_alignment}
\end{figure}

\Cref{fig:qualitative} illustrates representative predictions from the validation set.  
Our aerial-guided model reconstructs intersection layouts and road boundaries with improved geometric continuity,
exhibiting fewer false positives and sharper lane geometry compared to the baseline.  
Notably, in the top side of the shown example, the baseline hallucinates a spurious boundary that is correctly suppressed when using aerial supervision.  
Similarly, the divider count is more accurately matched, approaching the quality of the fusion-based teacher model.

These improvements are reflected in the learned feature space.  
As shown in \Cref{fig:bev_alignment}, the \gls{bev} representations produced by the guided model exhibit clearer structural activations 
compared to the baseline encoder, particularly in regions with limited or occluded sensor input.  
For visualization, \gls{bev} feature maps are obtained by computing channel-wise mean activations that are normalized with a shared global scale to highlight spatial structure.  
The aerial encoder’s structured \gls{bev} features extracted from the top-down view act as dense geometric priors, 
encouraging more coherent and topologically consistent outputs.

\section{Discussion and Limitations}
\label{sec:discussion}

Camera-based \gls{bev} perception relies on ego-centric observations, which can make maintaining globally consistent spatial representations difficult at larger distances.
Our results suggest that perspective-privileged supervision helps address this limitation by reshaping the learning dynamics of ego-centric \gls{bev} encoders.
Rather than merely correcting local prediction errors, the aerial signal encourages globally consistent geometric reasoning, particularly in distant regions where ego-centric observations provide limited spatial evidence.  
Structural information from complementary viewpoints therefore acts as a stabilizing prior for long-range spatial understanding.
This behavior suggests that aerial supervision provides dense structural guidance for \gls{bev} feature representations that is not accessible through supervision based on discretized semantic map targets.

In contrast to LiDAR–camera distillation or diffusion-based refinement, which operate entirely within the ego-centric sensing domain, our approach leverages supervision from a viewpoint whose structural information remains stable across large spatial extents.  
The overhead perspective directly exposes road layout and connectivity, providing geometric cues that are largely unaffected by occlusions, sparsity, or perspective distortion.  
As a result, aerial guidance promotes smoother \gls{bev} representations and more coherent road geometries, while leaving the inference architecture and runtime unchanged.

The main practical limitation of CVS is the need for accurately ego-aligned aerial imagery during training. 
While high-resolution overhead imagery is increasingly available through public orthophotos~\cite{DOP20_BKG, OS_UK, BDORTHO_FR}, commercial satellite imagery, or targeted drone captures, the critical requirement is precise metric registration to the local ego-centric coordinate frame. 
AID4AD demonstrates a semi-automatic high-precision alignment workflow for this purpose, providing a practical basis for cross-view supervision. 
Further automating and scaling this alignment process is an important step toward applying CVS across larger datasets and deployment regions.
Beyond alignment, Appendix~\ref{app:weather} shows that CVS remains effective under ego-view weather variation, reducing relative rainy-scene degradation from 21.2\% to 10.0\%.

These requirements also constrain the scope of the empirical evaluation.
Because AID4AD currently provides cross-view aligned aerial imagery for nuScenes, our experiments focus on this benchmark.
Extending high-quality cross-view alignment to additional datasets such as Argoverse~2~\cite{Argoverse2} would enable broader validation across sensor setups, geographic environments, and traffic layouts.

Future work should further investigate cross-view supervision across additional tasks and model architectures, including occupancy prediction, motion forecasting, and broader structured scene understanding.

\section{Conclusion}
\label{sec:conclusion}

We introduced cross-view supervision as a representation learning paradigm that transfers structural priors from a perspective-privileged overhead view into ego-centric \gls{bev} encoders for online \gls{hd} map construction.  
By bridging complementary viewpoints during training, the proposed framework enables camera-only models to internalize global geometric structure without requiring additional sensors or inference overhead.

Our experiments on nuScenes with the AID4AD dataset demonstrate that perspective-privileged supervision consistently improves spatial consistency in learned \gls{bev} representations, with particularly strong gains at extended spatial ranges where ego-centric observations provide limited structural evidence.  
These results indicate that complementary viewpoints can act as effective supervisory signals for shaping the geometry of \gls{bev} feature spaces rather than merely refining downstream predictions.

More broadly, this work highlights cross-view supervision as a promising training mechanism for camera-based spatial perception.
Unlike conventional auxiliary objectives that supervise \gls{bev} encoders through semantic map projections, our approach enables dense feature-level supervision using perspective-aligned aerial imagery.
Incorporating such perspective-privileged signals during training may provide a promising direction for enhancing the global reasoning capabilities of \gls{bev}-based models, with potential applications extending beyond map construction to other structured scene understanding tasks.

\vspace{0.15em}
\noindent\textbf{Acknowledgment}
This work is supported by the NeMo.bil project 19S23003, which
is funded by the Federal Ministry for Economic Affairs and Energy of Germany.

\bibliographystyle{splncs04}
\bibliography{main}

\clearpage
\appendix
\section*{Appendix}

This document provides additional details, analyses, and visualizations that complement the main paper.

\section{Sensitivity to the BEV loss weight $\lambda_{\text{bev}}$}

\begin{figure}
    \centering
    \includegraphics[width=0.8\linewidth]{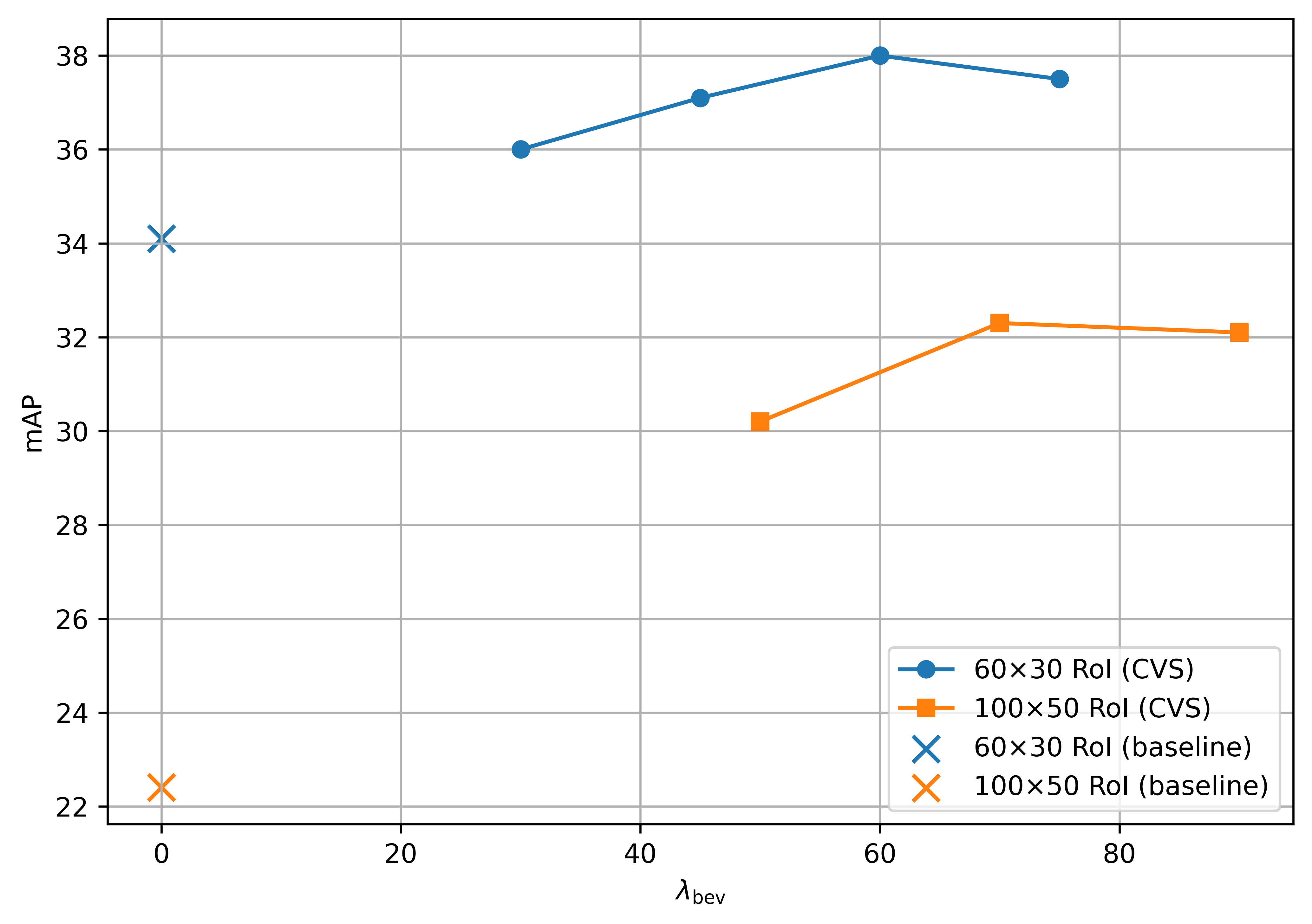}
    \caption{
    Sensitivity of performance to the cross-view supervision weight $\lambda_{\text{bev}}$.
    Results remain stable across a wide range of values, indicating low sensitivity to the exact loss weight.
    }
    \label{fig:bev_lambda_sensitivity}
\end{figure}

We analyze the effect of different BEV loss weights for two regions of
interest (60×30 m and 100×50 m). The points at
$\lambda_{\text{bev}}=0$ correspond to models without BEV supervision,
while all $\lambda_{\text{bev}}>0$ represent the supervised setting.
As shown in \Cref{fig:bev_lambda_sensitivity}, performance increases
sharply when adding any BEV supervision, after which further changes
in $\lambda_{\text{bev}}$ have only minor impact. This indicates that
the method is robust to the exact choice of this weight.

The observed variations are notably smaller than the gains from the
cross-view projection components in \Cref{tab:adapter_ablation}.
Enabling normalization and the affine adapter improves performance by
up to +9.9 mAP, clearly exceeding the fluctuations caused by
different $\lambda_{\text{bev}}$ values. This confirms that
cross-view alignment is the dominant performance factor.

\section{Additional Qualitative BEV Feature Visualizations}

\begin{figure}
  \centering
  \setlength{\tabcolsep}{3pt}
  \renewcommand{\arraystretch}{1.0}
  \includegraphics[width=0.7\linewidth]{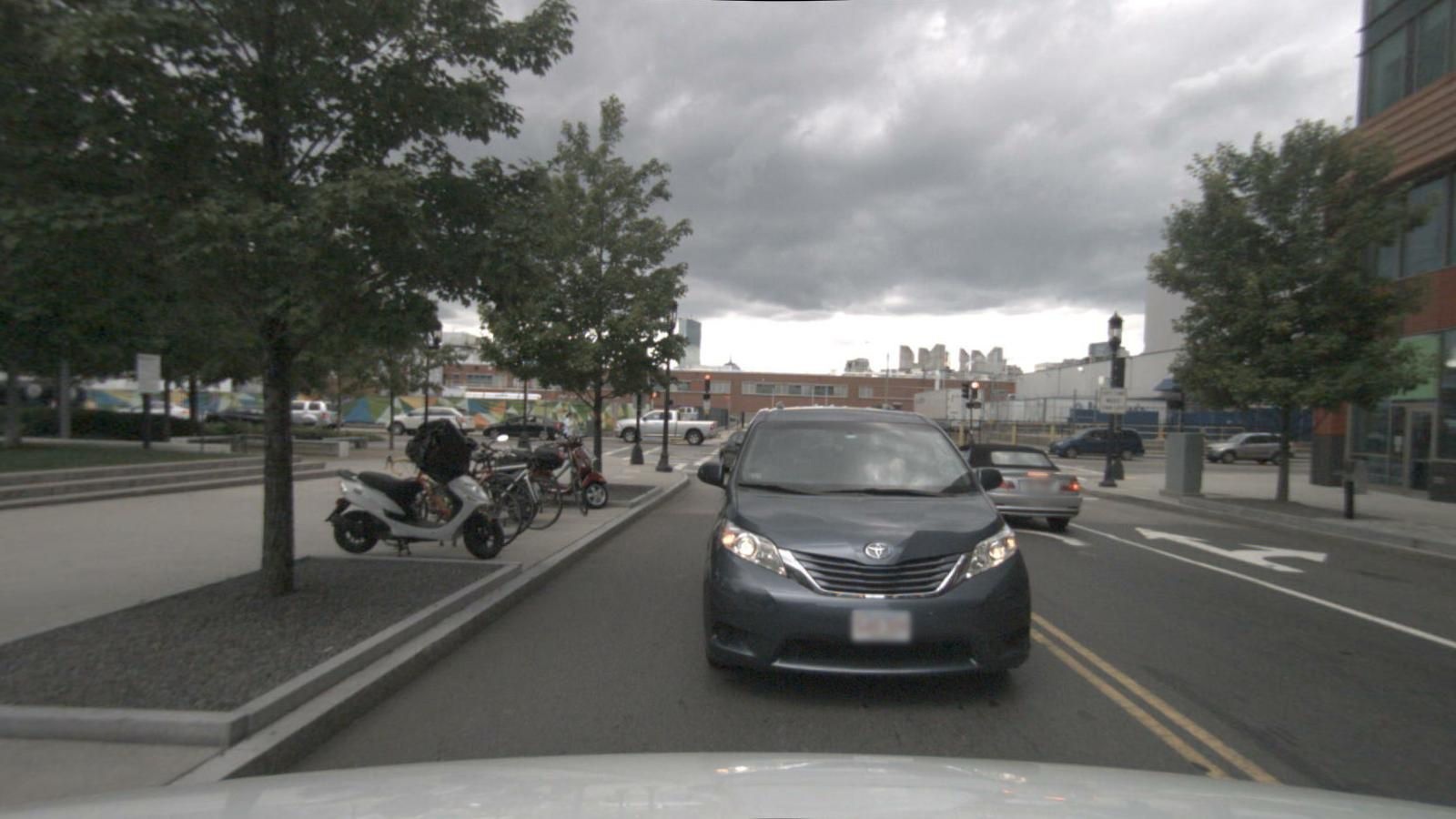}\\[0.3em]
  \small Back Camera View \\[0.8em]

  \begin{tabular}{cc}
    \includegraphics[width=0.33\linewidth]{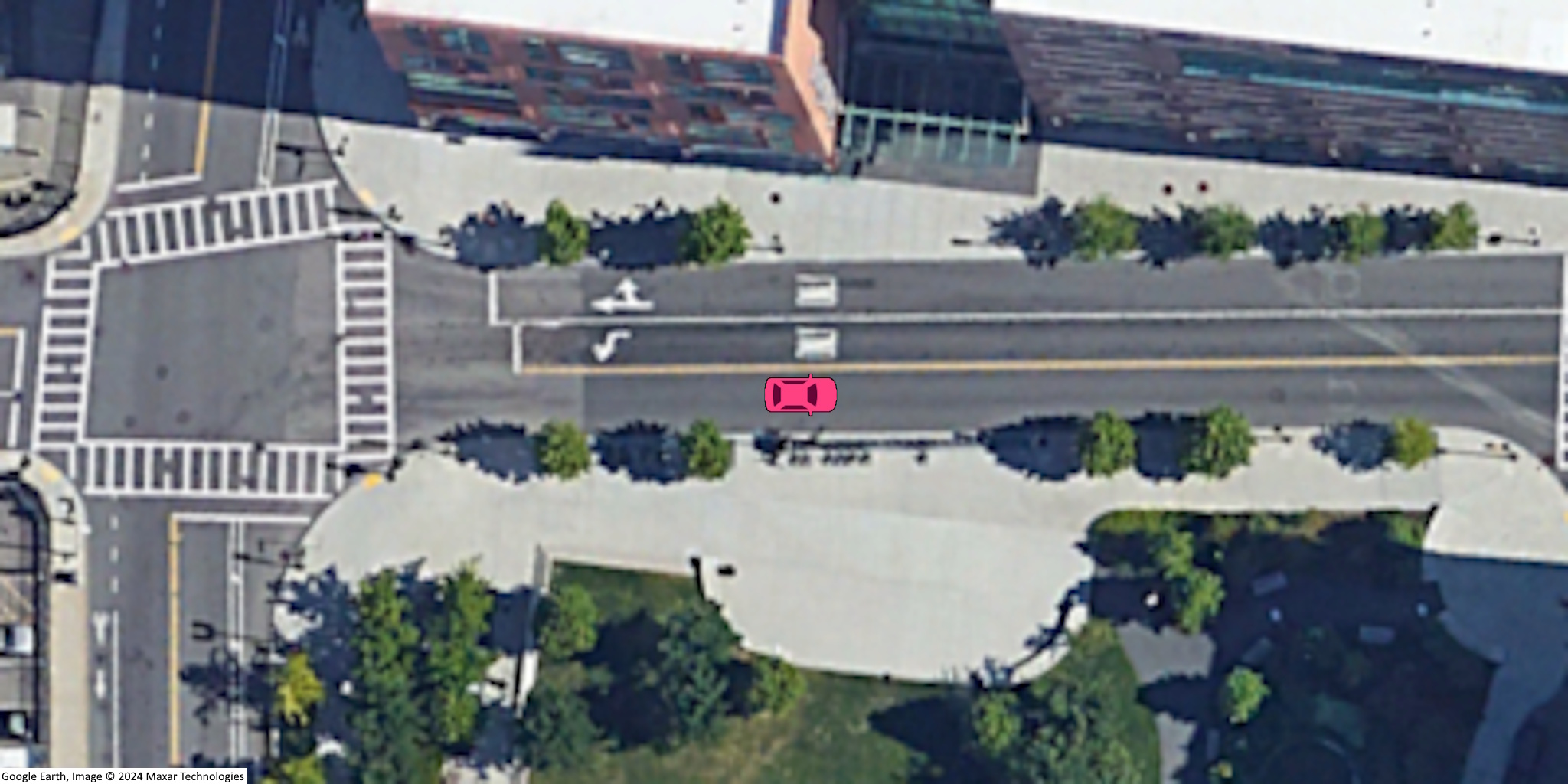} &
    \includegraphics[width=0.33\linewidth]{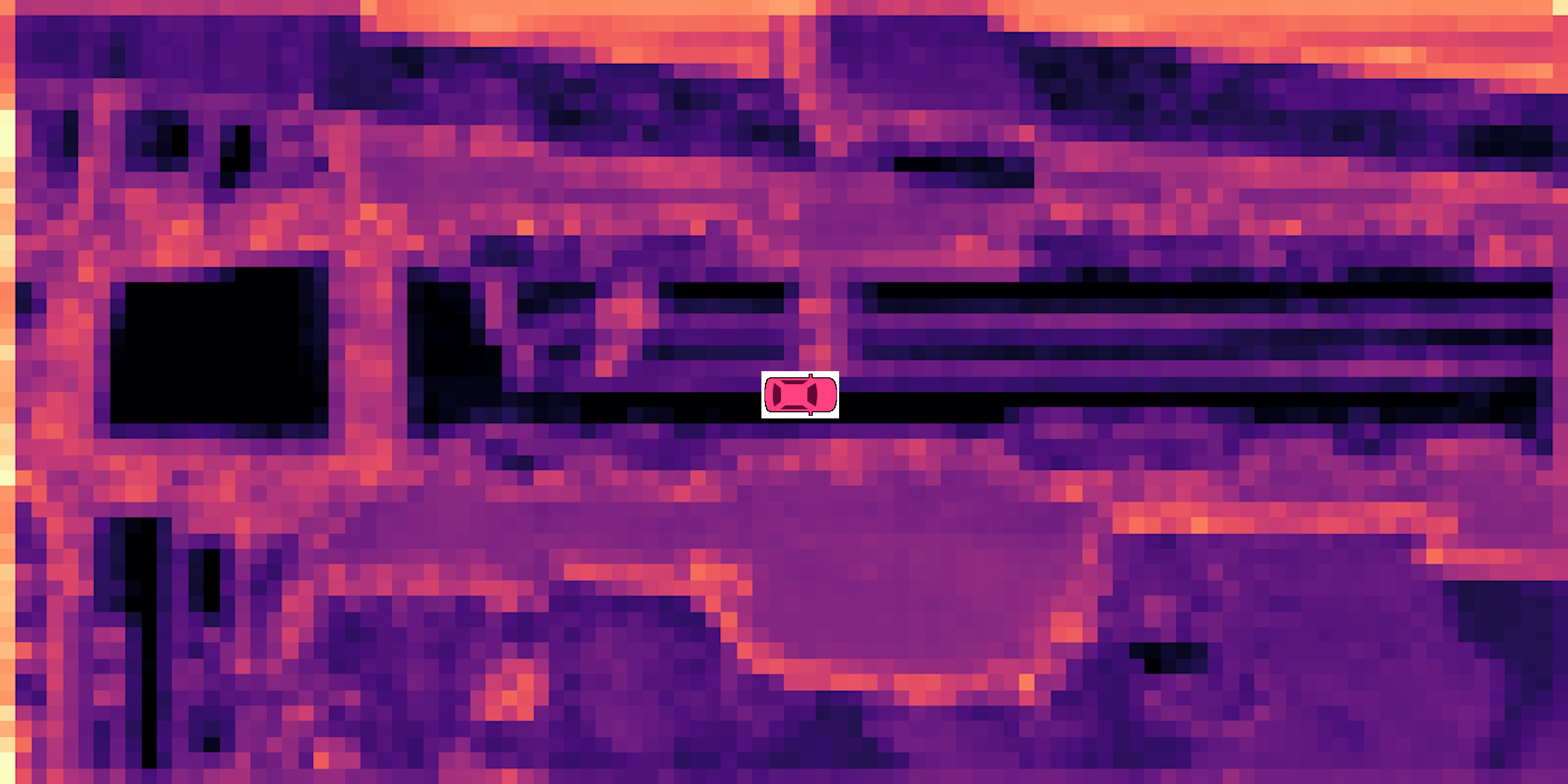} \\
    \small Aerial RGB & \small Aerial BEV (teacher) \\[0.6em]

    \includegraphics[width=0.33\linewidth]{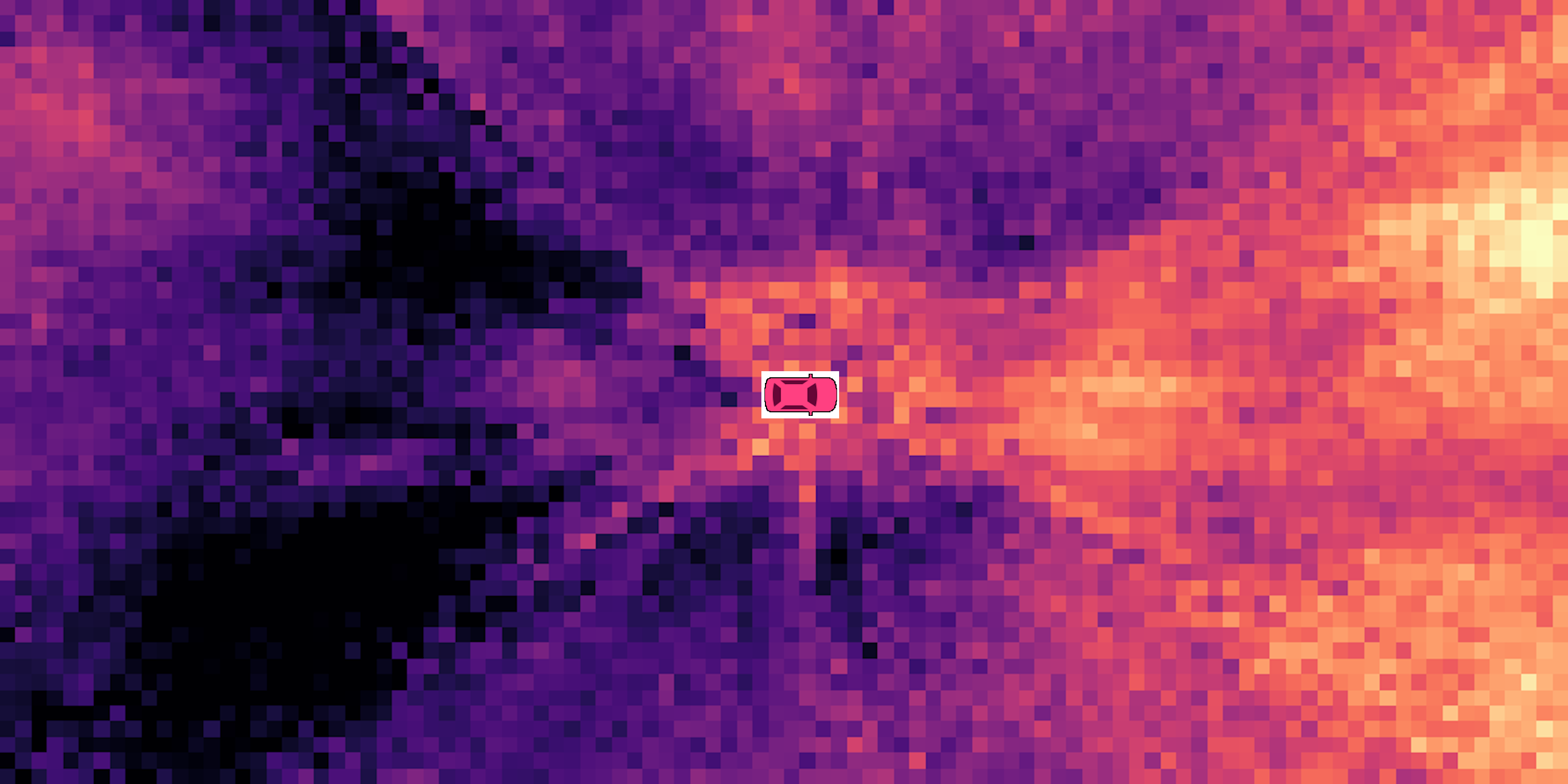} &
    \includegraphics[width=0.33\linewidth]{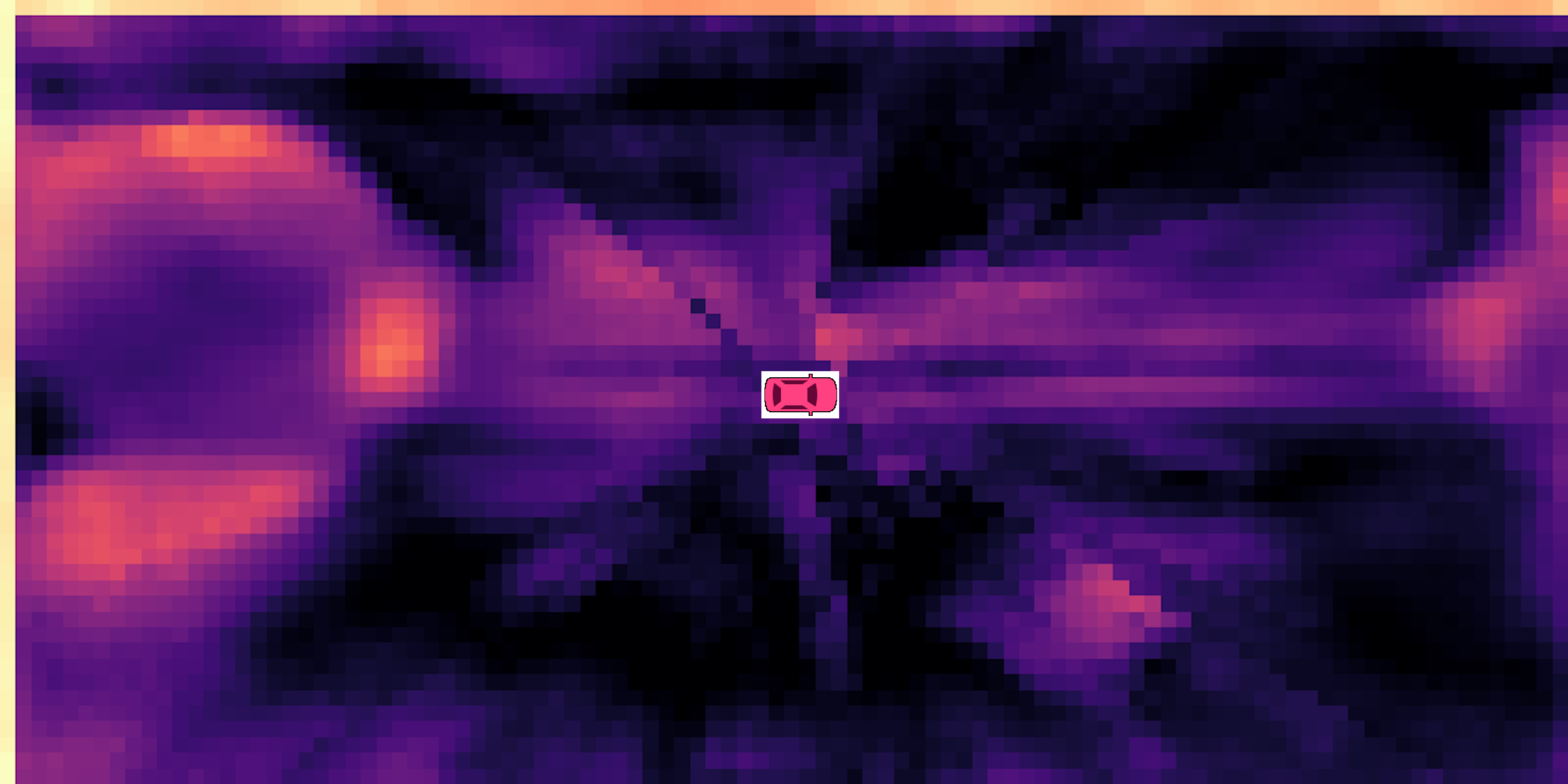} \\
    \small Camera BEV (baseline) & \small Camera BEV (ours)
  \end{tabular}
  \caption{
  Sample~1 illustrates an occlusion case where vehicles block the road and only faint hints of pedestrian crossings remain. The supervised features recover the road boundaries more clearly than the baseline and align well with the aerial teacher.
    }
  \label{fig:sample1}
\end{figure}

To further illustrate the effect of aerial feature supervision on the learned
representations, we present additional BEV feature maps.  
Each sample includes (i) the aligned aerial RGB patch from AID4AD,
(ii) the BEV feature map of the aerial teacher,
(iii) the camera-only baseline, and (iv) our cross-view guided model.
For \Cref{fig:sample1}, we additionally show the corresponding back camera view to
highlight an occlusion case.

Across both examples, the guided model produces more structured and spatially
coherent activations that better capture lane topology, curb geometry, and road
boundaries. In contrast, the baseline often yields diffuse or noisy responses,
especially under occlusion or limited camera coverage, confirming that aerial
supervision provides a strong structural prior for improved spatial consistency.

\begin{figure}
  \centering
  \setlength{\tabcolsep}{3pt}
  \renewcommand{\arraystretch}{1.0}

  \begin{tabular}{cc}
    \includegraphics[width=0.33\linewidth]{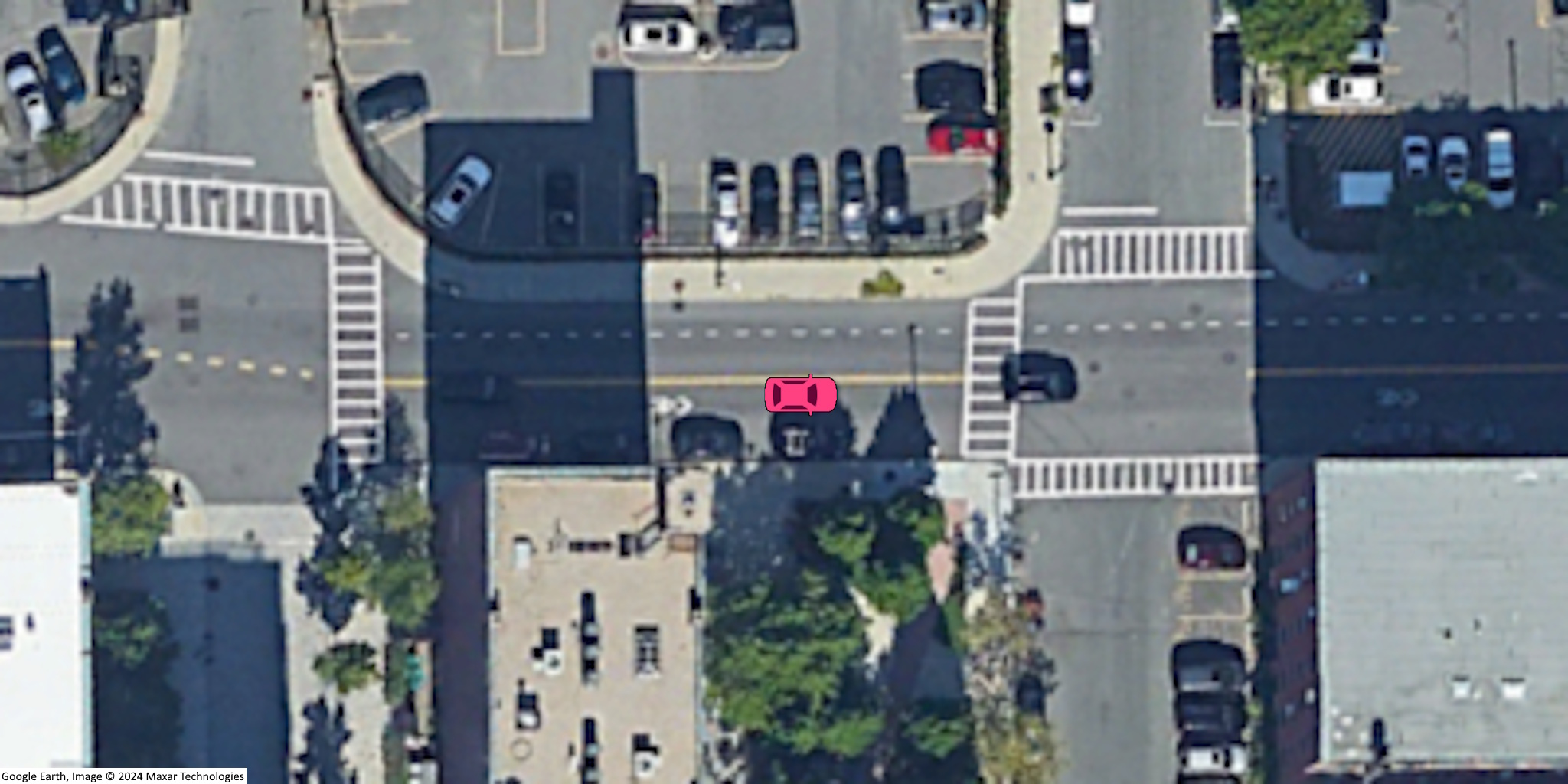} &
    \includegraphics[width=0.33\linewidth]{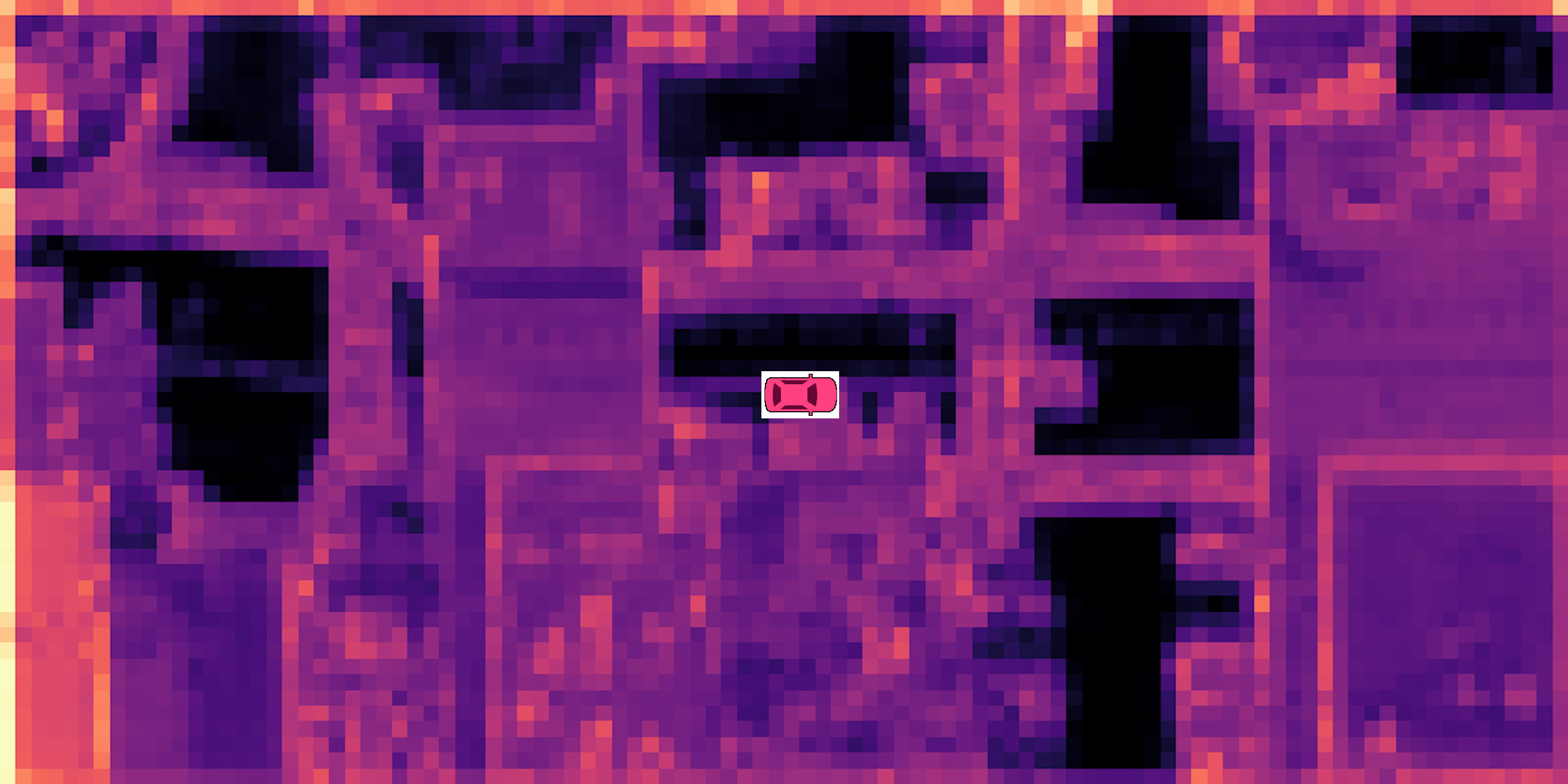} \\
    \small Aerial RGB & \small Aerial BEV (teacher) \\[0.6em]

    \includegraphics[width=0.33\linewidth]{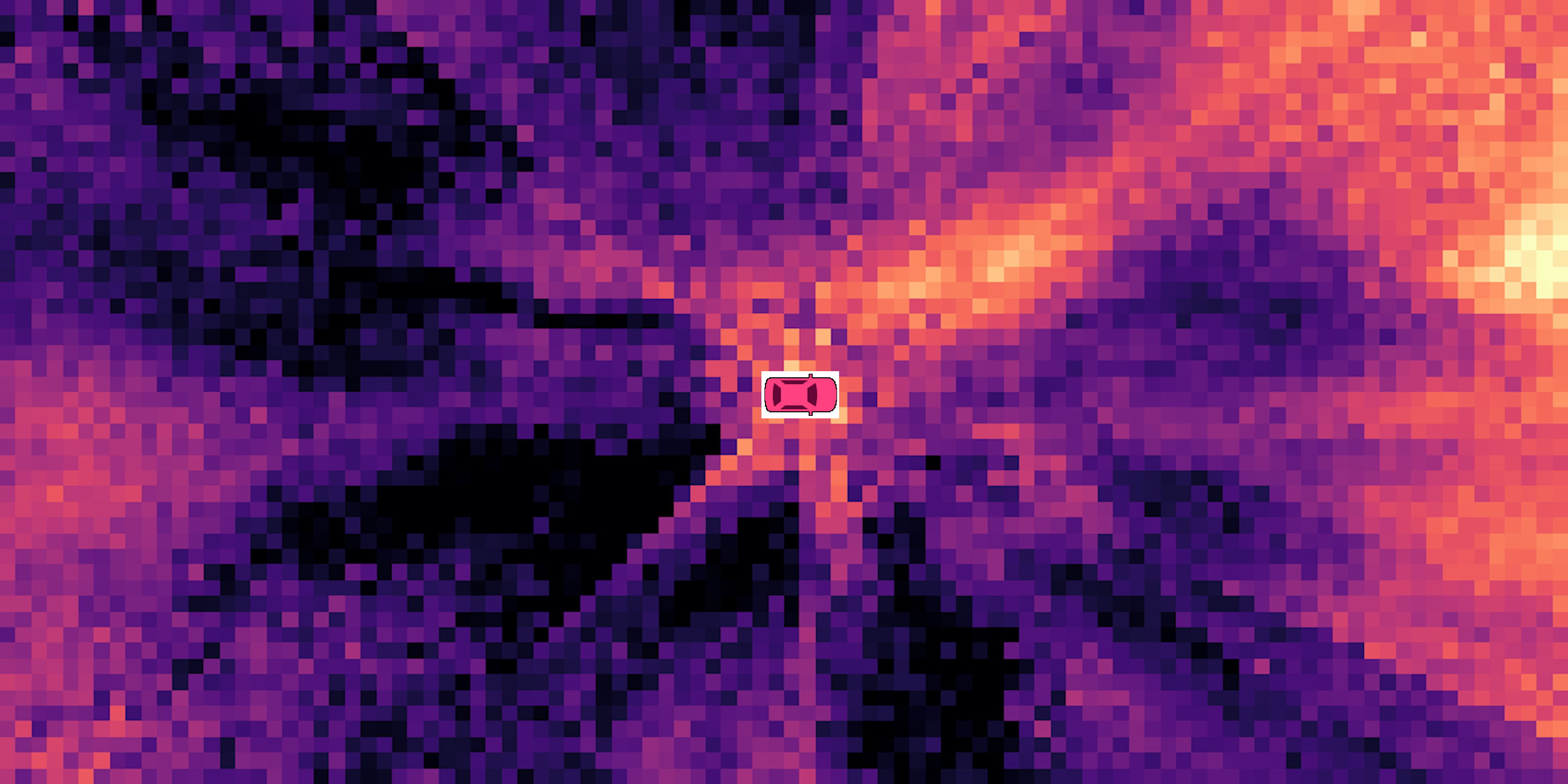} &
    \includegraphics[width=0.33\linewidth]{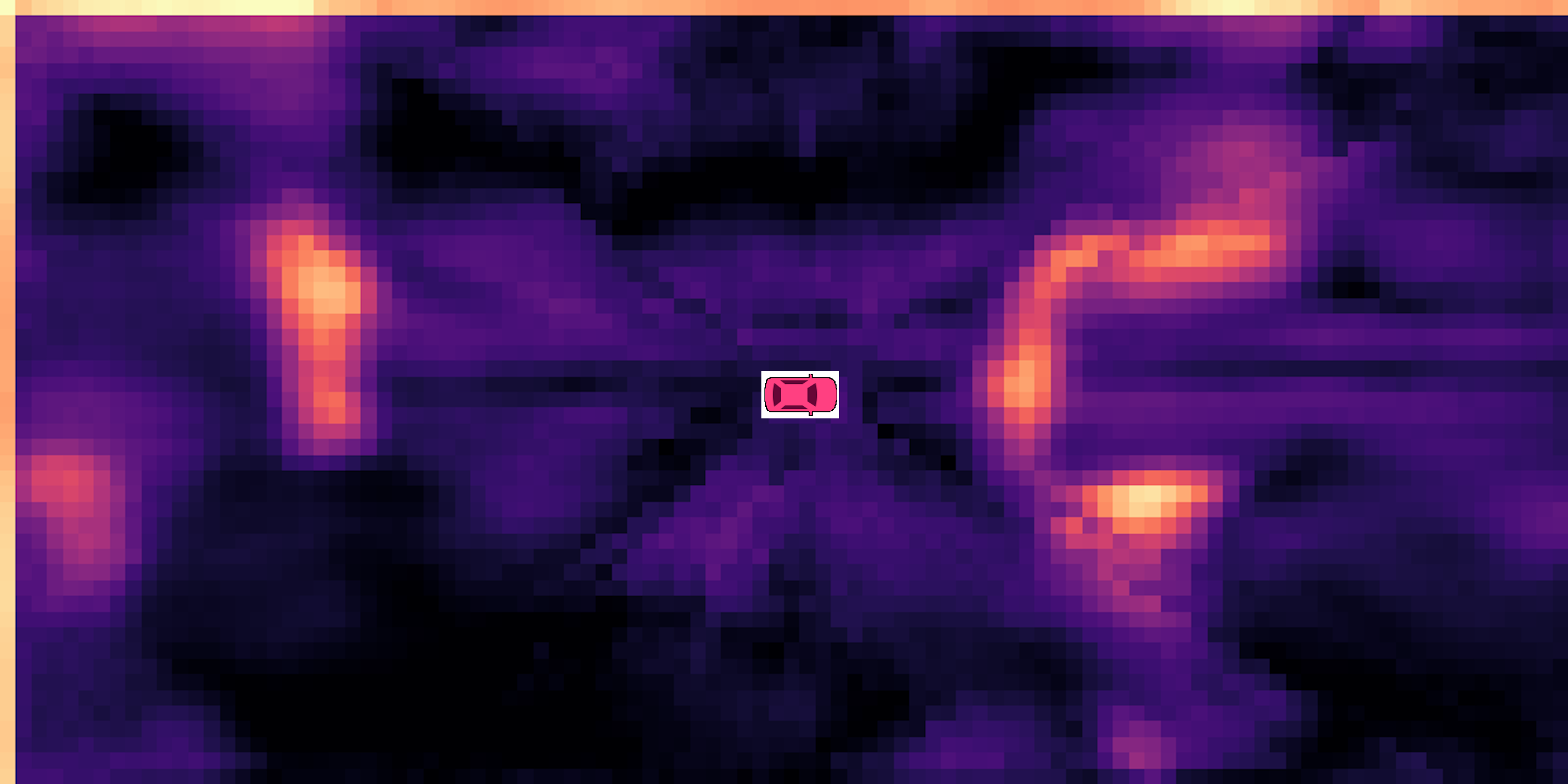} \\
    \small Camera BEV (baseline) & \small Camera BEV (ours)
  \end{tabular}

  \caption{
    Sample~2 shows that aerial supervision yields sharper and more stable road
    geometry than the baseline, consistent with the aerial teacher features.
  }
\end{figure}

\section{Teacher Feature Transferability}
\label{app:teacher_features}

\begin{figure}
  \centering
  \setlength{\tabcolsep}{3pt}
  \renewcommand{\arraystretch}{1.0}

  \begin{tabular}{cc}
    \includegraphics[width=0.33\linewidth]{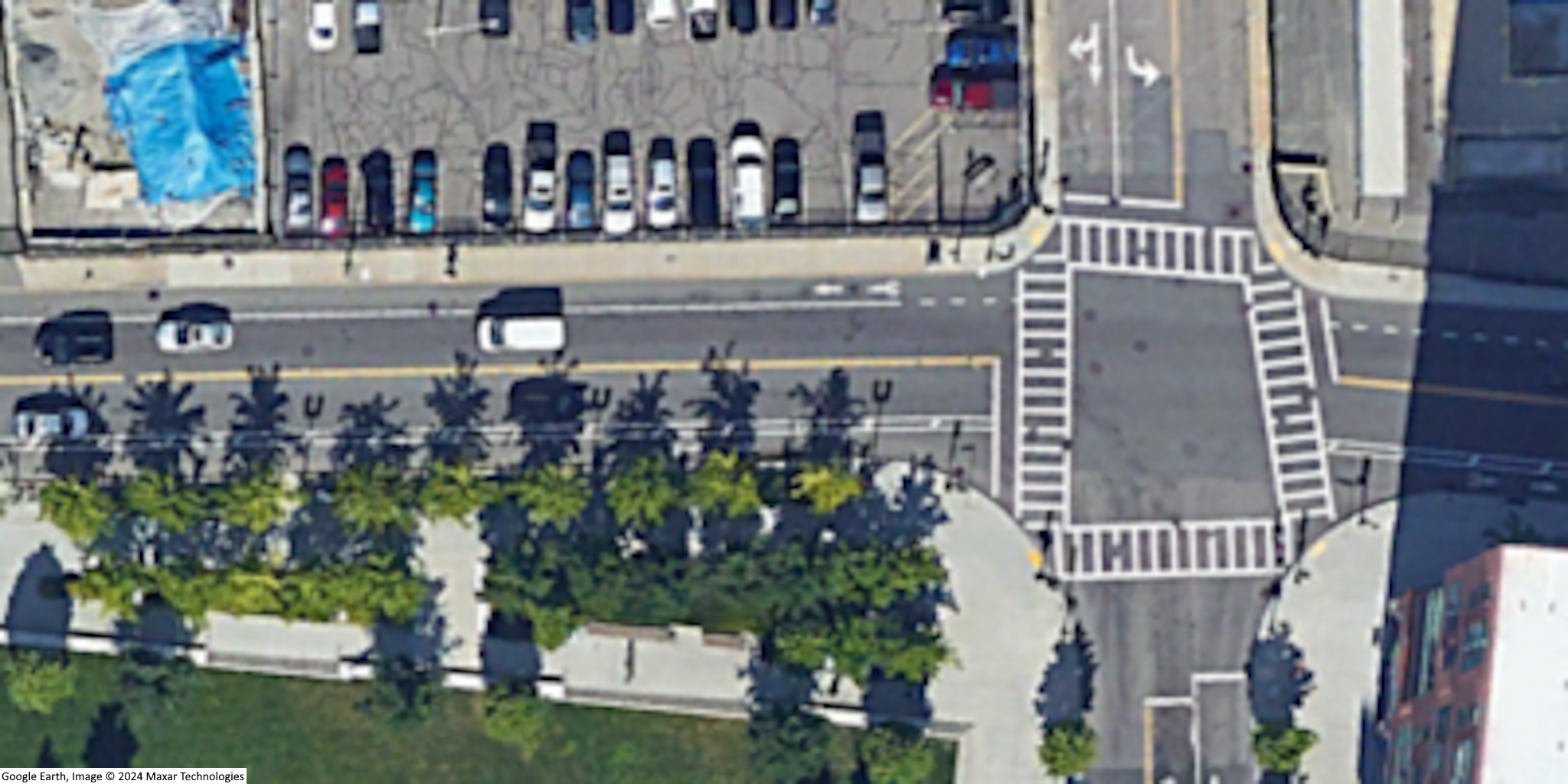} &
    \includegraphics[width=0.33\linewidth]{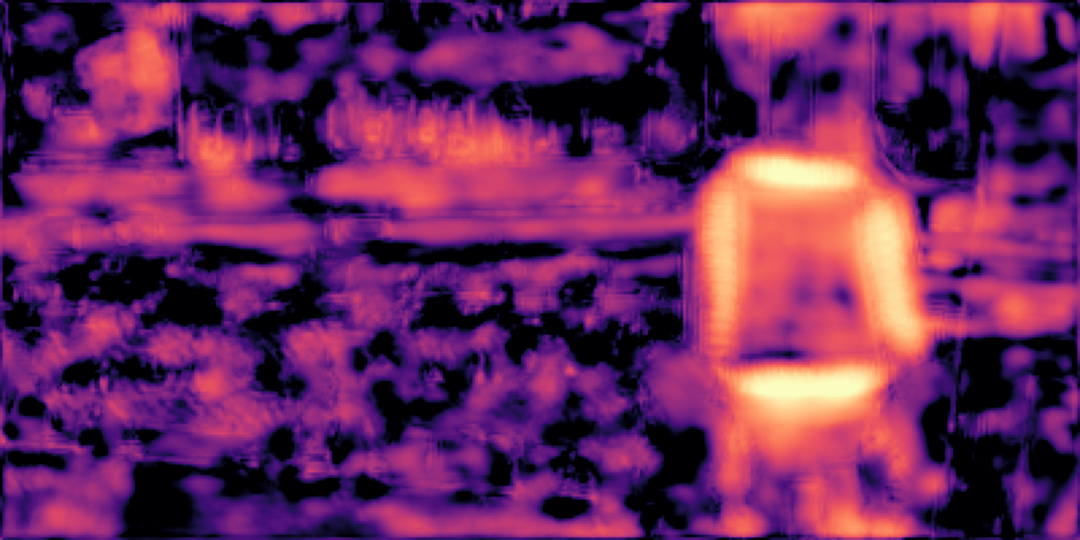} \\
    \small Aerial RGB & \small ResUNet \\[0.6em]

    \includegraphics[width=0.33\linewidth]{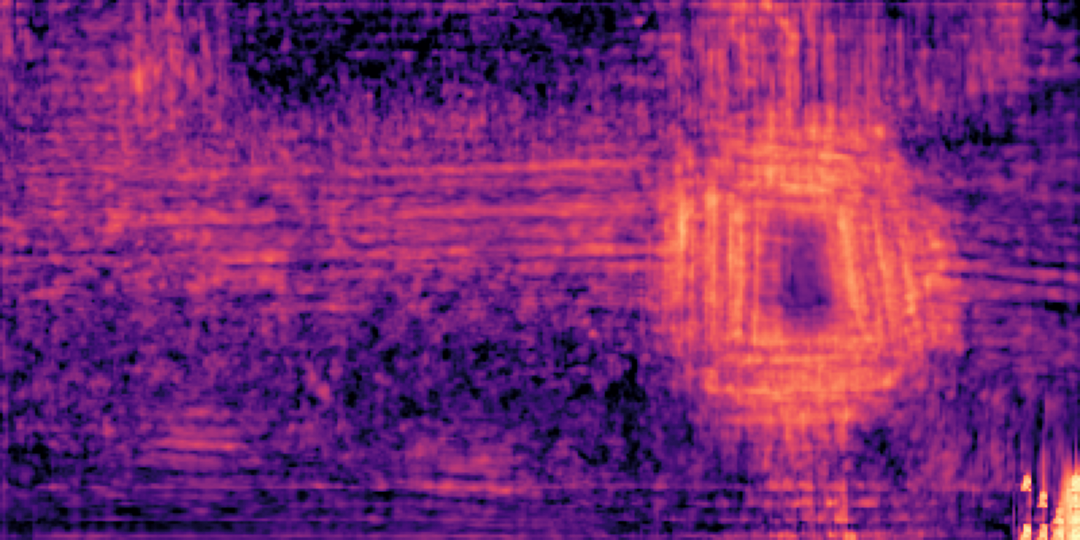} &
    \includegraphics[width=0.33\linewidth]{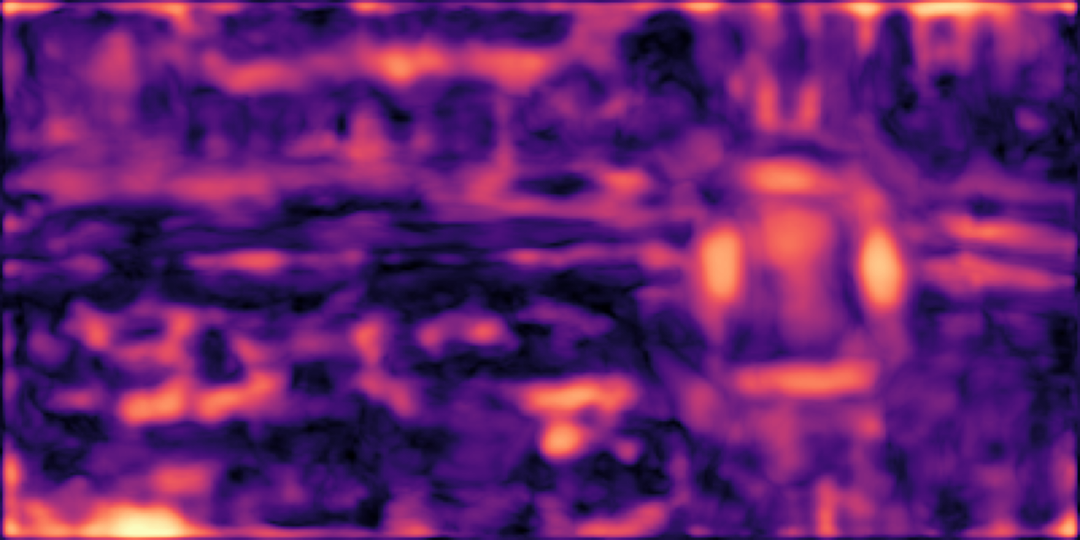} \\
    \small ResUNet++ & \small UNet++
  \end{tabular}

  \caption{
    Aerial image and teacher feature activations for the same scene/crop.
    Feature maps are visualized as normalized channel-wise mean activations
    with shared scaling.
  }
\label{fig:teacher_features_app}
\end{figure}

To further analyze the influence of teacher architecture, Fig.~\ref{fig:teacher_features_app} visualizes teacher feature activations for the same aerial crop.
All maps are shown as normalized channel-wise mean activations with shared scaling.
Compared to ResUNet and UNet++, ResUNet++ produces less localized and higher-frequency activation patterns.
This supports the interpretation that teacher-side mapping accuracy alone does not determine CVS performance; the intermediate representation must also be suitable as a dense feature-level target for the camera-based BEV encoder.

\section{Weather-Stratified Evaluation}
\label{app:weather}

To assess whether CVS remains effective under ego-view weather variation, we stratify the validation set into rainy and non-rainy scenes without retraining.
Table~\ref{tab:weather} shows that CVS improves both subsets and reduces the relative rainy-scene degradation from 21.2\% to 10.0\%.

\begin{table}[h]
\centering
\small
\caption{Weather-stratified validation results on the \(100\times50~\mathrm{m}\) setting.}
\label{tab:weather}
\begin{tabular}{lccc}
\toprule
Subset & \# samples & StreamMapNet & CVS \\
\midrule
Non-rainy & 4763 & 23.6 & 33.0 \\
Rainy & 1218 & 18.6 & 29.7 \\
\midrule
Rainy degradation & -- & 21.2\% & 10.0\% \\
\bottomrule
\end{tabular}
\end{table}

\end{document}